\documentclass{article}

   \usepackage[preprint, nonatbib]{neurips_2023}

\usepackage[utf8]{inputenc} % allow utf-8 input
\usepackage[T1]{fontenc}    % use 8-bit T1 fonts
\usepackage{hyperref}       % hyperlinks
\usepackage{url}            % simple URL typesetting
\usepackage{booktabs}       % professional-quality tables
\usepackage{amsfonts}       % blackboard math symbols
\usepackage{nicefrac}       % compact symbols for 1/2, etc.
\usepackage{microtype}      % microtypography
\usepackage{xcolor}         % colors
\usepackage{float}
\usepackage{amsmath, amsthm, amsfonts,amscd, amssymb, bm}
\usepackage{multicol}
\usepackage{multirow}
\usepackage{enumitem}
\usepackage{graphicx}
\usepackage{caption}
\usepackage{algorithm}
\usepackage{algorithmic}
\usepackage{bbm}
\usepackage{pifont}
\usepackage{moresize}
\usepackage{array}

\newcolumntype{C}[1]{>{\centering\arraybackslash}b{#1}}

\title{Saliency–Depth Conditioning for Zero-Shot Segmentation of Communication-Tower Components in Cluttered UAV Imagery}

\author{%
  Ali Lesani$^{1,2}$, Chul Min Yeum$^{1,2}$, Su-Min Kang$^{3}$ 
  \\[2ex]
  \small $^{1}$ Computer Vision for Smart Structures (CViSS) Lab, Waterloo, Canada \\
  \small $^{2}$ University of Waterloo, Waterloo, Canada \\
  \small $^{3}$ School of Architecture, Soongsil University, Seoul, South Korea
}

\begin{document}

\maketitle

\begin{abstract}
Fine-grained segmentation of communication-tower components in UAV imagery is essential for automated inspection, yet task-specific models are hard to develop due to limited instance-level annotations. Zero-shot segmentation models offer a promising alternative, but in cluttered scenes, visually similar background structures interfere with component localization, causing missed instances and false positives. We propose a model-agnostic saliency--depth foreground-conditioning strategy combining appearance-based saliency with monocular relative depth to construct a coarse tower prior and suppress irrelevant content. We integrate this module with Grounded-SAM and SAM~3, yielding SD-Grounded-SAM and SD-SAM~3. SD-Grounded-SAM further applies geometric and depth-aware box refinement before mask generation, while SD-SAM~3 relies on SAM~3's internal setup. On TOW-300, a dataset of 340 communication-tower UAV images, our strategy improves both baselines: SD-SAM~3 achieves the strongest instance-segmentation performance, while SD-Grounded-SAM produces fewer false positives. Ablations confirm complementary gains from saliency, depth, and box refinement, improving robustness in cluttered scenes. 
\end{abstract}

% //////////////////////////////////////////////////////////////////
% /////////////////////////  Introduction  /////////////////////////
% //////////////////////////////////////////////////////////////////
\section{Introduction}
Communication towers are critical components of modern wireless infrastructure, enabling large-scale cellular connectivity. Accurate visual understanding of these structures---particularly the ability to identify and segment individual components such as antennas and radios---is essential for downstream applications including asset inventory, automated maintenance, change detection, structural assessment, anomaly detection, and network optimization \cite{lyu2025uav, faisal2025deep, villarino2025uav}. As these tasks increasingly rely on high-resolution visual data collected in the field, unmanned aerial vehicle (UAV)-based inspection has emerged as a practical solution, offering a safe, efficient, and cost-effective alternative to manual climbing or ground-based imaging. UAVs can capture comprehensive visual coverage of towers from multiple viewpoints and elevations, leading to widespread adoption in infrastructure monitoring \cite{lyu2025uav, faisal2025deep, li2024deep, wang2024deep}. However, the resulting images are often visually complex and contain substantial background clutter, making automated component-level segmentation challenging.

This challenge is further amplified by variations introduced during UAV image acquisition and by the visual characteristics of tower-mounted components. UAV flight paths produce considerable changes in viewpoint, scale, elevation, and perspective, causing the same component to appear differently across images. In addition, antennas and radio units are often small relative to the full-resolution image, overlap with the tower structure, or are partially occluded by neighboring equipment. Their localization is further complicated by surrounding vegetation, buildings, cables, power lines, and metallic or rectangular structures that may exhibit similar visual patterns. Related difficulties have also been reported in UAV-based inspection of power-transmission infrastructure, particularly for detecting small components under cluttered backgrounds and large scale variations \cite{faisal2025deep, li2024deep, liu2020deep, nejadbougarintelligent, wang2024enhanced}. Consequently, successful component segmentation depends not only on accurate mask generation, but also on reliable localization of small target structures in visually crowded scenes.

Addressing these localization and segmentation challenges has traditionally relied on supervised instance segmentation models \cite{bolya2019yolact, he2017mask, wang2020solov2}. These models can achieve strong performance when sufficient task-specific annotations are available. However, constructing large instance-level datasets for communication-tower components is costly and labor-intensive because each small, overlapping, or partially occluded component must be annotated at the pixel level. Moreover, models trained on a fixed dataset may not generalize reliably to changes in tower configuration, antenna design, image-acquisition conditions, and surrounding environments. Data scarcity and distribution shift therefore remain major barriers to the practical deployment of supervised segmentation models in infrastructure inspection \cite{lyu2025uav, faisal2025deep, wang2024deep}. These limitations motivate the investigation of training-free approaches that leverage general-purpose pre-trained models without requiring a dedicated segmentation model for each tower configuration.

Recent foundation models provide new opportunities for prompt-based and zero-shot segmentation. The Segment Anything Model (SAM) can generate high-quality masks from user-provided prompts, such as points or bounding boxes, and has demonstrated strong transfer across diverse image distributions \cite{kirillov2023segment}. However, SAM does not independently localize objects of interest and therefore depends on accurate prompts to produce meaningful masks. In large-scale UAV inspection, manually specifying prompts for every antenna or radio unit is impractical. Grounded-SAM addresses this limitation by combining the open-vocabulary detector Grounding DINO with SAM, allowing bounding-box prompts to be generated automatically from natural-language descriptions \cite{liu2024grounding, ren2024grounded}. More recent concept-prompted models, such as SAM~3, further unify object localization and mask generation within a single zero-shot segmentation framework.

Despite these advances, open-vocabulary localization remains difficult in communication-tower imagery. Object names such as \textit{antenna} are ambiguous because the detector has limited domain-specific knowledge of tower-mounted equipment. In our observations, such prompts may identify the complete tower, structural members, or unrelated objects rather than individual mounted components. More descriptive appearance-based prompts can improve localization of antenna-like regions, but they also increase false detections on background elements with similar geometry. These errors arise during object localization and can propagate to the final masks, even when the underlying segmentation model is capable of accurately delineating a prompted region. Therefore, the central challenge is not only mask generation, but also reducing the visual ambiguity presented to the zero-shot model.

This limitation motivates the use of visual pre-processing before prompt-based detection and segmentation. In particular, reducing irrelevant background content can remove many non-target objects that satisfy coarse geometric text descriptions and thereby reduce ambiguity for downstream zero-shot models. To this end, we introduce a model-agnostic saliency--depth foreground-conditioning strategy for training-free segmentation of communication-tower components in UAV imagery. The proposed module combines appearance-based saliency with monocular relative depth to construct a coarse tower prior. Saliency-based background suppression first isolates visually prominent foreground regions and reduces the search space presented to subsequent models \cite{kim2022revisiting}. However, saliency estimation may remove low-contrast, partially occluded, or visually weak portions of the tower. We therefore incorporate monocular depth estimation, exploiting the observation that the communication tower is generally closer to the UAV camera than much of the surrounding background \cite{yang2024depth}. Depth cues are used to recover near-field regions that may have been omitted by saliency estimation. Morphological refinement and connected-component filtering are then applied to produce a coherent foreground prior, which is used to construct a conditioned image with reduced background interference.

We integrate the proposed conditioning strategy with two distinct zero-shot segmentation frameworks. First, we develop Saliency--Depth Grounded-SAM (SD-Grounded-SAM), a domain-aware extension of Grounded-SAM. In this configuration, Grounding DINO operates on the conditioned image to generate candidate boxes for tower-mounted components. The candidate detections are subsequently refined using tower-relative geometry, pairwise containment, and relative-depth cues. The refined boxes are then provided to SAM as spatial prompts for instance-mask generation on the original image. Second, we apply the same saliency--depth conditioning module to SAM~3, resulting in Saliency--Depth SAM~3 (SD-SAM~3). In contrast to Grounded-SAM, SAM~3 performs localization and segmentation internally; therefore, the conditioned image is passed directly to its concept-prompted inference pipeline without an explicit box-refinement stage. These two integrations allow us to evaluate whether the proposed foreground-conditioning strategy improves architecturally distinct zero-shot segmentation systems.

We evaluate the proposed method using TOW-300, a newly developed dataset comprising 340 high-resolution UAV images collected under real-world field conditions, with instance-level annotations of tower-mounted components. The saliency--depth conditioning strategy consistently improves both Grounded-SAM and SAM~3 across instance- and image-level segmentation metrics. The SD-SAM~3 configuration achieves the strongest overall instance segmentation performance, while the full SD-Grounded-SAM pipeline provides a more conservative operating point with higher precision and cleaner aggregate foreground masks.

We summarize our primary contributions as follows: (1) we introduce a model-agnostic and training-free saliency--depth foreground-conditioning strategy that combines complementary appearance and geometric cues to reduce background ambiguity in cluttered UAV imagery; (2) we integrate the proposed conditioning strategy with two distinct zero-shot segmentation frameworks, resulting in SD-Grounded-SAM and SD-SAM~3, and demonstrate consistent improvements over their corresponding unconditioned baselines; (3) for the Grounded-SAM-based integration, we develop an additional box-refinement stage that uses tower-relative geometry, pairwise containment, and relative-depth cues to produce more reliable spatial prompts for SAM; and (4) we conduct a comprehensive evaluation on real-world communication-tower imagery, showing that the proposed conditioning strategy improves instance localization, recall, mask overlap, and aggregate foreground quality across both segmentation frameworks.

% //////////////////////////////////////////////////////////////////
% /////////////////////////  Related Work  /////////////////////////
% //////////////////////////////////////////////////////////////////
\section{Related Work}
\paragraph{UAV-Based Visual Inspection.}
UAV-based visual inspection is widely used to collect visual data from infrastructure that is difficult, costly, or unsafe to access manually \cite{messaoudi2023survey,bouhamed2020uav,alfattani2019multi}. Recent reviews show that UAVs are used across a broad range of infrastructure applications, including bridges, buildings, transportation assets, and utility systems, because they enable flexible image acquisition and support efficient downstream analysis \cite{bouhamed2020uav, pomarnacki2025unmanned}. In this workflow, deep learning has become an important tool for automating the detection, classification, and segmentation of defects and structural components from UAV imagery \cite{lyu2025uav,korki2019automatic,farhangfar2019semantic,agyemang2023automated,xiao2024region}. In aerial infrastructure vision, closely related work has focused on transmission towers, power lines, and insulators \cite{sangaiah2025lcut,lan2025amfa,zhao2025pl,lan2025amfa}. For example, TTPLA introduced a benchmark dataset for detection and segmentation of transmission towers and power lines in aerial images \cite{abdelfattah2020ttpla}, while other studies have developed UAV-based methods for insulator detection under complex backgrounds and large scale variation \cite{si2024research,faisal2025deep}. These studies establish UAV inspection as a practical computer vision setting, but they mainly address power infrastructure or defect detection rather than fine-grained segmentation of communication-tower antennas and radios.

\paragraph{Supervised Segmentation.}
Supervised segmentation methods remain the dominant approach for dense visual understanding. Architectures such as U-Net and DeepLabv3+ have become standard baselines for semantic segmentation, while models such as Mask R-CNN, YOLACT, and SOLOv2 are widely used for instance segmentation across domain-specific vision tasks \cite{ronneberger2015u,chen2018encoder,he2017mask,bolya2019yolact,wang2020solov2,calabrese2025application,xu2022crack}. In infrastructure inspection, supervised models have also been adapted to aerial tower and power-line imagery, often with task-specific datasets and closed-set label definitions \cite{abdelfattah2020ttpla,zhu2023corner,si2024research}. While these methods can achieve strong results when sufficient labeled data are available, they are less attractive in settings where fine-grained annotation is expensive and target components vary across hardware types and deployment environments. Our work differs from this line by addressing communication-tower component segmentation in a zero-shot setting without training a dedicated supervised segmenter.

\paragraph{Foundation Models for Zero-Shot Segmentation.}
Recent foundation models have significantly advanced zero-shot and open-vocabulary segmentation by leveraging large-scale visual and vision--language pretraining. Methods such as CLIPSeg, LSeg, and GroupViT explore language-guided segmentation and open-vocabulary recognition across diverse image domains \cite{luddecke2022image,li2022language,xu2022groupvit}. SAM further introduced large-scale promptable segmentation and demonstrated strong transfer to previously unseen image distributions \cite{kirillov2023segment}. Building on SAM, Grounded-SAM combines open-vocabulary detection with promptable mask generation to enable text-guided instance segmentation \cite{liu2024grounding,ren2024grounded}, while more recent models such as SAM~3 integrate concept-based localization, segmentation, and tracking within a unified framework \cite{carion2025sam}. Despite their broad generalization capabilities, these models are designed as general-purpose vision systems rather than being tailored to the structural characteristics of infrastructure-inspection imagery \cite{zhou2024teaching,pu2025classwise}. Existing approaches primarily improve segmentation through stronger prompting, open-vocabulary localization, or unified detection--segmentation architectures, whereas comparatively less attention has been given to modifying the visual input itself to reduce domain-specific scene ambiguity. Our work addresses this gap by introducing a model-agnostic foreground-conditioning stage based on complementary saliency and depth cues. Rather than modifying or fine-tuning the underlying zero-shot models, the proposed strategy restructures the visual evidence presented to them, allowing both Grounded-SAM and SAM~3 to operate on a reduced and more relevant search space.

% //////////////////////////////////////////////////////////////////
% /////////////////////////    Method    ///////////////////////////
% //////////////////////////////////////////////////////////////////
\section{Method}
\subsection{Problem Setup}
In this work, we address the problem of zero-shot instance-level segmentation of communication-tower components from UAV-captured RGB images. Given an input image
$\mathbf{I}\!\in\!\mathbb{R}^{H\times W\times 3}$, our goal is to generate a set of instance masks
\begin{equation}
\mathcal{M}
=
\left\{
\mathbf{M}_1,
\mathbf{M}_2,
\dots,
\mathbf{M}_K
\right\},
\label{eq:masks}
\end{equation}
where $K$ denotes the number of predicted component instances, and
$\mathbf{M}_k \in \{0,1\}^{H \times W}$ denotes the binary mask of the $k$th predicted instance, with $k \in \{1,\dots,K\}$. Each mask corresponds to an individual tower-mounted component, such as an antenna or radio unit.

The objective is to produce accurate instance-level masks without task-specific model training. Instead of training a dedicated segmentation network, we use general-purpose pre-trained models and introduce a shared saliency--depth foreground-conditioning module that reduces background ambiguity before zero-shot segmentation. Given the input image, the conditioning module produces a refined foreground image
\begin{equation}
\mathbf{I}_{\mathrm{ref}}
=
\mathcal{R}(\mathbf{I}),
\label{eq:conditioning_function}
\end{equation}
where $\mathcal{R}(\cdot)$ denotes the proposed saliency--depth conditioning operation. The resulting conditioned image $\mathbf{I}_{\mathrm{ref}}$, together with a textual prompt $\mathbf{p}$ describing the target components, is then provided to one of two zero-shot segmentation frameworks:
\begin{equation}
\mathcal{M}^{(j)}
=
f_j
\left(
\mathbf{I},
\mathbf{I}_{\mathrm{ref}},
\mathbf{p}
\right),
\qquad
j
\in
\left\{
\mathrm{GSAM},
\mathrm{SAM3}
\right\},
\label{eq:segmentation_frameworks}
\end{equation}
where $f_{\mathrm{GSAM}}(\cdot)$ denotes the SD-Grounded-SAM pipeline and
$f_{\mathrm{SAM3}}(\cdot)$ denotes the SD-SAM~3 pipeline.

\begin{figure}
    \centering
    \includegraphics[width=\linewidth]{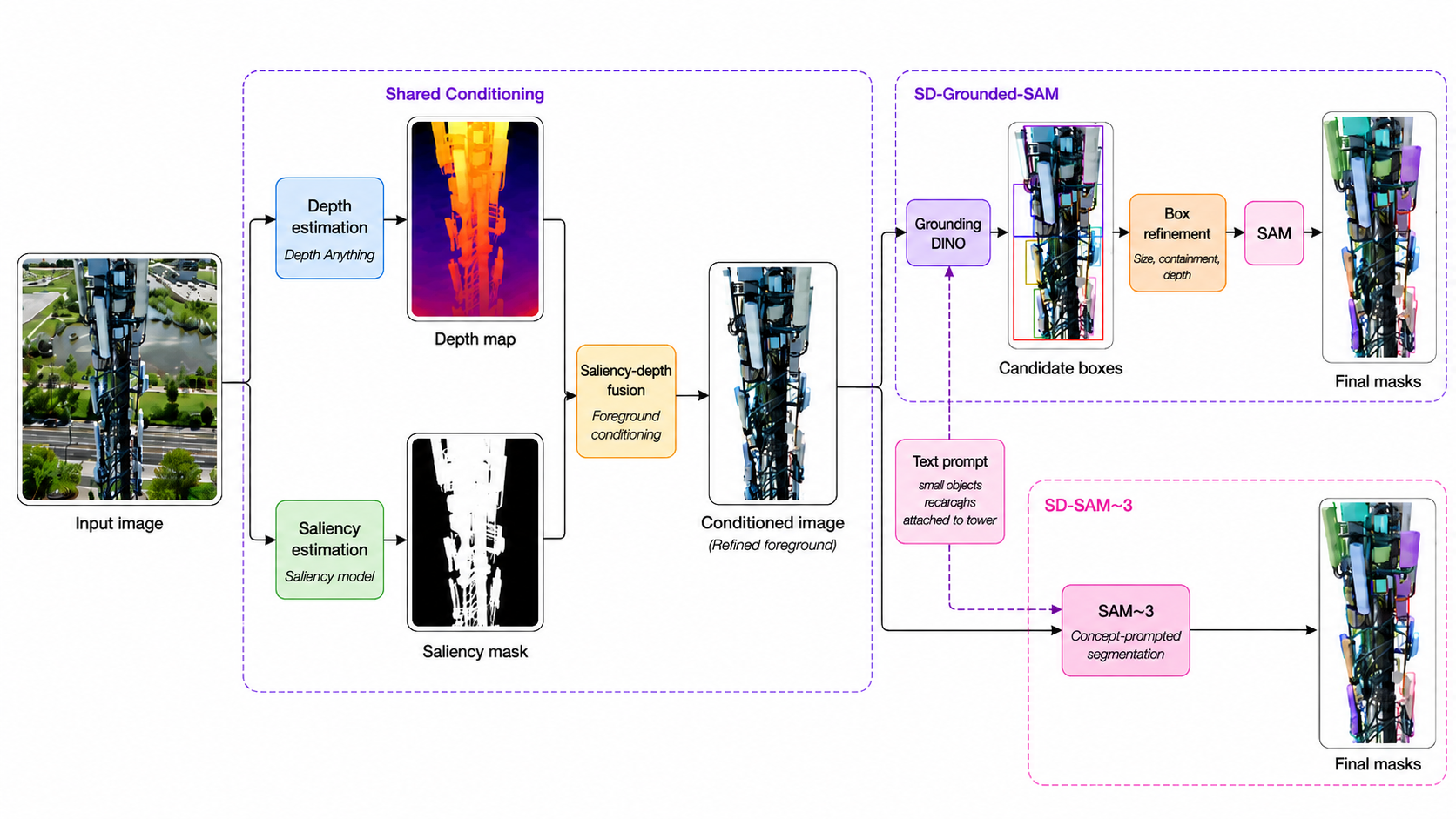}
    \caption{\textbf{Overview of the proposed saliency--depth conditioning framework.} Given an input UAV image, saliency estimation and monocular depth prediction are combined to construct a refined foreground mask and a conditioned image with reduced background interference. The conditioned image is then processed by two zero-shot segmentation frameworks. In SD-Grounded-SAM, Grounding DINO generates candidate boxes that are refined using tower-relative geometry, pairwise containment, and depth cues before being passed to SAM. In SD-SAM~3, the conditioned image and text prompt are provided directly to SAM~3, which performs localization and segmentation internally.}
    \label{fig:pipeline}
\end{figure}

\subsection{Saliency--Depth Foreground Conditioning}

An overview of the proposed framework is shown in Figure~\ref{fig:pipeline}. Starting from a raw RGB image captured by a UAV, the conditioning module first estimates visually salient foreground regions and predicts a monocular relative-depth map. The two cues are then combined to construct a coarse tower prior. Morphological refinement and connected-component filtering are applied to remove isolated noise and obtain a coherent foreground region. The resulting mask is used to generate a conditioned image that preserves the dominant tower structure while suppressing much of the surrounding background. The conditioning module is shared by both downstream zero-shot segmentation frameworks. Its purpose is not to perform final component segmentation directly, but to reduce the visual ambiguity presented to the segmentation model.

\paragraph{Saliency-Based Foreground Estimation.}

To reduce the effect of background clutter, we first estimate a saliency map from the input image. We use the Transparent Background model, which is based on a deep salient-object-detection network that separates visually prominent foreground regions from the background \cite{kim2022revisiting}. Given an input image
$\mathbf{I}\!\in\!\mathbb{R}^{H\times W\times 3}$,
the model produces a saliency map
\begin{equation}
\mathbf{S}
\in
[0,1]^{H \times W},
\label{eq:saliency}
\end{equation}
where larger values indicate a higher likelihood of belonging to the salient foreground. The saliency map is converted into an initial binary foreground mask:
\begin{equation}
\mathbf{M}_{\mathrm{sal}}(x,y)
=
\mathbbm{1}
\left[
\mathbf{S}(x,y)
\geq
\tau_{\mathrm{sal}}
\right],
\label{eq:saliency_mask}
\end{equation}
where $\tau_{\mathrm{sal}}$ is the saliency threshold and
$\mathbbm{1}[\cdot]$ denotes the indicator function. Applying this mask suppresses large portions of the irrelevant background and produces an initial foreground estimate focused primarily on the tower structure. However, saliency-based methods rely mainly on appearance cues and may fail under low contrast, occlusion, or complex textures. As a result, portions of the tower may be incorrectly suppressed as background. We therefore refine the foreground estimate using monocular depth information.

\paragraph{Depth-Based Foreground Refinement.}

To recover relevant regions that may be removed during saliency estimation, we incorporate monocular depth prediction using Depth Anything \cite{yang2024depth}. Given the same input image $\mathbf{I}$, the model produces a relative depth map
\begin{equation}
\mathbf{D}
\in
\mathbb{R}^{H \times W}.
\label{eq:depth}
\end{equation}

Based on the observation that the communication tower and its mounted components are typically closer to the UAV camera than much of the surrounding background, the depth map provides a complementary geometric cue for foreground identification. We construct a binary near-field recovery mask as
\begin{equation}
\mathbf{M}_{\mathrm{dep}}(x,y)
=
\mathbbm{1}
\left[
\mathbf{D}(x,y)
\geq
\tau_{\mathrm{rec}}
\right],
\label{eq:depth_recovery_mask}
\end{equation}
where $\tau_{\mathrm{rec}}\in[0,255]$ is the depth threshold used to identify near-field pixels. In the employed depth representation, larger values correspond to regions estimated to be closer to the camera.

\paragraph{Saliency--Depth Mask Fusion and Foreground Construction.}

The saliency and depth masks are first combined through a pixel-wise union:
\begin{equation}
\mathbf{M}_{0}
=
\mathbf{M}_{\mathrm{sal}}
\lor
\mathbf{M}_{\mathrm{dep}},
\label{eq:initial_foreground_mask}
\end{equation}
where $\lor$ denotes the pixel-wise logical OR operation. This operation preserves regions identified as salient while reintroducing near-field regions that may have been incorrectly removed by the saliency model.

The combined mask may still contain small holes, disconnected regions, and isolated noise. We therefore refine it using morphological closing followed by morphological opening:
\begin{equation}
\widetilde{\mathbf{M}}
=
\operatorname{Open}_{\mathcal{K}_{5}}
\left(
\operatorname{Close}_{\mathcal{K}_{9}}
\left(
\mathbf{M}_{0}
\right)
\right),
\label{eq:morphological_refinement}
\end{equation}
where $\mathcal{K}_{9}$ and $\mathcal{K}_{5}$ denote
$9\times9$ and $5\times5$ square structuring elements, respectively. Morphological closing fills small gaps and connects nearby foreground regions, whereas morphological opening removes isolated noise and small spurious structures.

We then retain the largest connected component in the morphologically refined mask:
\begin{equation}
\mathbf{M}_{\mathrm{ref}}
=
\operatorname{LCC}_{8}
\left(
\widetilde{\mathbf{M}}
\right),
\label{eq:refined_foreground_mask}
\end{equation}
where $\operatorname{LCC}_{8}(\cdot)$ denotes selection of the largest connected component using $8$-neighborhood connectivity. This step removes remaining disconnected regions under the assumption that the tower forms the dominant foreground structure.

The conditioned foreground image is obtained by applying the resulting mask to the original RGB image:
\begin{equation}
\mathbf{I}_{\mathrm{ref}}
=
\mathbf{I}
\odot
\mathbf{M}_{\mathrm{ref}},
\label{eq:refined_foreground_image}
\end{equation}
where $\odot$ denotes element-wise multiplication and $\mathbf{M}_{\mathrm{ref}}$ is broadcast across the three image channels. The resulting image retains the dominant salient and near-field structure while suppressing much of the surrounding background, thereby reducing the visual ambiguity presented to the downstream zero-shot segmentation model.

\subsection{Integration with Zero-Shot Segmentation Frameworks}

The proposed conditioning module is independent of the downstream segmentation architecture. We integrate it with Grounded-SAM and SAM~3 to evaluate whether reducing background ambiguity benefits zero-shot systems with different localization and segmentation mechanisms.

\subsubsection{SD-Grounded-SAM}

SD-Grounded-SAM is a domain-aware extension of Grounded-SAM. In this configuration, Grounding DINO operates on the conditioned image
$\mathbf{I}_{\mathrm{ref}}$
to generate candidate boxes for tower-mounted components. The candidate detections are subsequently refined using tower-relative geometry, pairwise containment, and relative-depth cues. The refined boxes are then provided to SAM as spatial prompts for instance-mask generation on the original image.

\paragraph{Open-Vocabulary Detection for Box-Prompt Generation.}

To automatically generate bounding-box prompts, we employ Grounding DINO \cite{liu2024grounding}, an open-vocabulary object detector that localizes objects from natural-language queries. The conditioned foreground image
$\mathbf{I}_{\mathrm{ref}}$
is provided to Grounding DINO together with the text prompt $\mathbf{p}$ describing the target tower-mounted components. The detector produces a set of candidate bounding boxes
\begin{equation}
\mathcal{B}
=
\left\{
\mathbf{B}_1,
\mathbf{B}_2,
\ldots,
\mathbf{B}_N
\right\}.
\label{eq:boxes}
\end{equation}

Although foreground conditioning suppresses much of the irrelevant scene content, the detector may still produce false-positive, redundant, or overly broad candidate boxes. The detections are therefore passed to a subsequent box-refinement stage before being used as spatial prompts for SAM.

\paragraph{Bounding-Box Post-processing.}

To improve the quality of the spatial prompts supplied to SAM, we apply a structured post-processing procedure based on box size, pairwise containment, and relative depth. First, we remove detections that are disproportionately large relative to the estimated tower region. To define this reference region, we compute the tight bounding box enclosing the final refined foreground mask
$\mathbf{M}_{\mathrm{ref}}$.
Because individual tower-mounted components are expected to be substantially smaller than the tower structure itself, candidate boxes whose width or height exceeds a fraction
$\tau_s \in [0,1]$
of the corresponding tower bounding-box dimension are discarded. These detections typically correspond to the entire tower or to large structural regions rather than individual antennas or radio units.

Next, we remove redundant or nested detections using an asymmetric pairwise containment measure. For two boxes
$\mathbf{B}_i$ and $\mathbf{B}_j$, we define
\begin{equation}
C(\mathbf{B}_i,\mathbf{B}_j)
=
\frac{
\left|
\mathbf{B}_i \cap \mathbf{B}_j
\right|
}{
\left|
\mathbf{B}_i
\right|
},
\label{eq:containment_ratio}
\end{equation}
where $|\cdot|$ denotes box area. Unlike standard Intersection over Union (IoU), this measure quantifies the proportion of
$\mathbf{B}_i$
contained within
$\mathbf{B}_j$
and is therefore suitable for identifying nested detections of different sizes. If
\begin{equation}
C(\mathbf{B}_i,\mathbf{B}_j)
>
\tau_{\mathrm{iou}},
\label{eq:containment_filter}
\end{equation}
the lower-confidence or otherwise redundant box is removed, where
$\tau_{\mathrm{iou}}\in[0,1]$
is the containment threshold.

Finally, we use the relative-depth map to suppress detections located farther from the camera. For each remaining box
$\mathbf{B}_i$,
the mean depth value is computed over the pixels inside the box:
\begin{equation}
\mu_i
=
\frac{1}{|\Omega_i|}
\sum_{(x,y)\in\Omega_i}
\mathbf{D}(x,y),
\label{eq:mean_box_depth}
\end{equation}
where $\Omega_i$ denotes the set of pixels enclosed by
$\mathbf{B}_i$.
Boxes whose mean depth is classified as background according to the threshold
$\tau_d\in[0,255]$
are discarded.

This depth-based filtering is particularly useful in images captured from elevated or near top-down viewpoints, where rooftops, ground objects, and other background structures may exhibit shapes similar to tower-mounted components and may remain after foreground conditioning. The complete post-processing operation produces the refined box set
\begin{equation}
\mathcal{B}_{\mathrm{ref}}
=
\mathcal{P}
\left(
\mathcal{B},
\mathbf{D};
\tau_s,
\tau_{\mathrm{iou}},
\tau_d
\right),
\label{eq:box_postprocessing}
\end{equation}
where $\mathcal{P}(\cdot)$ denotes the box-refinement procedure.

\paragraph{Prompt-Based Segmentation.}

We use the SAM \cite{kirillov2023segment} to generate instance-level masks from the refined bounding boxes. Each refined bounding box
$\mathbf{B}_i \in \mathcal{B}_{\mathrm{ref}}$
is provided to SAM as a spatial prompt together with the original RGB image $\mathbf{I}$:
\begin{equation}
\mathbf{M}_i^{\mathrm{GSAM}}
=
\operatorname{SAM}
\left(
\mathbf{I},
\mathbf{B}_i
\right),
\label{eq:sam}
\end{equation}
where
$\mathbf{M}_i^{\mathrm{GSAM}}
\in
\{0,1\}^{H\times W}$
denotes the predicted binary mask for the $i$th retained component. Applying SAM to all refined boxes produces the final SD-Grounded-SAM output:
\begin{equation}
\mathcal{M}_{\mathrm{GSAM}}
=
\left\{
\mathbf{M}_1^{\mathrm{GSAM}},
\mathbf{M}_2^{\mathrm{GSAM}},
\ldots,
\mathbf{M}_{K_{\mathrm{G}}}^{\mathrm{GSAM}}
\right\},
\label{eq:final_gsam_masks}
\end{equation}
where
$K_{\mathrm{G}}
=
|\mathcal{B}_{\mathrm{ref}}|$
is the number of retained detections after post-processing. Although box generation is performed on the conditioned image, mask prediction is applied to the original RGB image to preserve the full visual detail of each component.

\subsubsection{SD-SAM~3}

We also integrate the same saliency--depth conditioning module with SAM~3, resulting in SD-SAM~3. Unlike the Grounded-SAM-based pipeline, SAM~3 performs concept localization and instance segmentation internally and therefore does not require externally generated bounding-box prompts. The conditioned foreground image
$\mathbf{I}_{\mathrm{ref}}$
and the same textual prompt $\mathbf{p}$ are supplied directly to the SAM~3 concept-prompted inference pipeline:
\begin{equation}
\mathcal{M}_{\mathrm{SAM3}}
=
\operatorname{SAM3}
\left(
\mathbf{I}_{\mathrm{ref}},
\mathbf{p}
\right),
\label{eq:sd_sam3}
\end{equation}
where
\begin{equation}
\mathcal{M}_{\mathrm{SAM3}}
=
\left\{
\mathbf{M}_1^{\mathrm{SAM3}},
\mathbf{M}_2^{\mathrm{SAM3}},
\ldots,
\mathbf{M}_{K_{\mathrm{S}}}^{\mathrm{SAM3}}
\right\}
\label{eq:sam3_masks}
\end{equation}
denotes the set of instance masks generated by SAM~3. No external box-generation or box-refinement stage is applied in this branch. Instead, the conditioning module modifies the visual input presented to SAM~3, while concept localization, instance identification, and mask generation remain internal to the model. This integration allows us to isolate the effect of saliency--depth foreground conditioning on a unified zero-shot segmentation architecture and to assess whether its benefit extends beyond the Grounded-SAM pipeline.

\subsection{Model Modularity}
Although the experiments in this work use Transparent Background for saliency estimation, Depth Anything for monocular depth prediction, Grounding DINO for open-vocabulary localization, and SAM/SAM 3 for segmentation, the proposed framework is not tied to these specific models. Each component can be replaced by an alternative model that produces a compatible intermediate output, such as a saliency map, relative-depth map, candidate bounding boxes, or instance masks. The main contribution therefore lies in the saliency--depth conditioning and integration strategy rather than in any particular model choice. The use of two distinct downstream segmentation frameworks further demonstrates this modularity.

% //////////////////////////////////////////////////////////////////
% /////////////////////////    Dataset    ///////////////////////////
% //////////////////////////////////////////////////////////////////

\section{Dataset}

We construct TOW-300, a UAV-based dataset for instance-level segmentation of communication-tower components. The dataset is collected from three separate UAV flights covering two visually and structurally distinct communication towers and contains 340 high-resolution images with multiple mounted components, primarily antennas and radio units. Each visible component is treated as an individual instance. The images have resolutions of either $5000 \times 6300$ or $9100 \times 6300$ pixels, preserving the fine-grained visual details required to delineate small and partially occluded components.

\paragraph{Data Collection.}
The images were collected during real-world tower inspections using a DJI Mavic 3 Enterprise UAV equipped with a 20 MP camera. \cite{dji_website}. To capture variation representative of practical inspection conditions, the images were acquired at different times of day and under varying illumination conditions. The UAV followed multiple flight trajectories, including helical paths around the tower and vertical lawnmower patterns, producing observations from different viewpoints, elevations, and distances. Consequently, the dataset exhibits substantial variation in component scale, perspective, occlusion, and background composition.

\begin{figure}[t]
    \centering
    \setlength{\tabcolsep}{2pt}
    \begin{tabular}{cc}
        \includegraphics[width=0.35\textwidth]{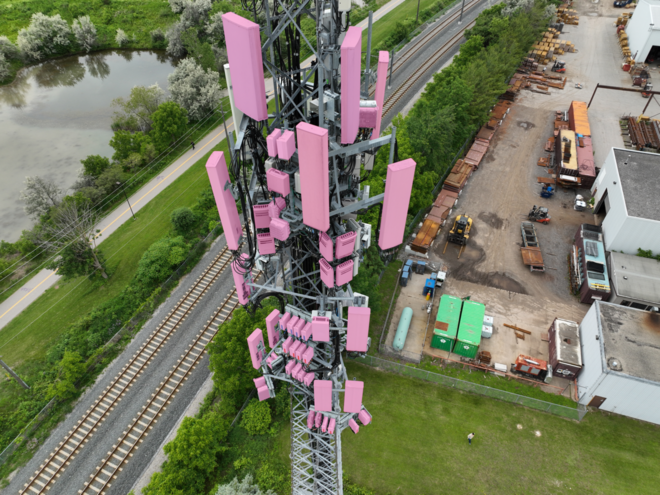} &
        \includegraphics[width=0.35\textwidth]{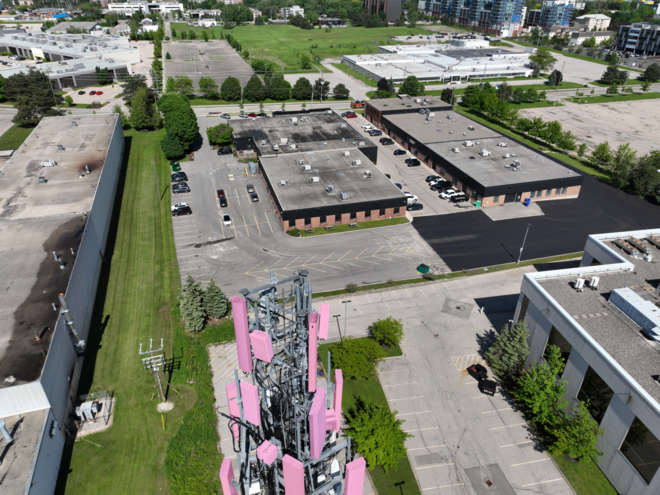} \\
        
        (a) & (b)
    \end{tabular}

    \caption{\textbf{Representative instance-level annotations from TOW-300.}
    All component instances are shown using the same overlay color (pink) for improved visibility and visual clarity.}
    \label{fig:sample-ann}
\end{figure}

\paragraph{Annotation.}
All images were annotated using the CVAT platform \cite{cvat_github}. Instance-level polygon masks were manually drawn around each visible tower-mounted component and exported in COCO format, including both segmentation masks and their corresponding bounding boxes \cite{lin2014microsoft}. The dataset contains a single semantic class, \textit{tower-mounted component}, which includes antennas and radio units, while preserving a separate annotation for each component instance. Partially visible components were annotated when their boundaries could be identified with sufficient confidence. Components for which only a small and highly ambiguous portion was visible were excluded to avoid introducing unreliable annotations. This policy was applied consistently throughout the dataset. Representative ground-truth annotations are shown in Figure~\ref{fig:sample-ann}. 

\begin{figure}[t]
    \centering
    \setlength{\tabcolsep}{2pt}
        \begin{tabular}{cccc}
            \includegraphics[width=0.241\textwidth]{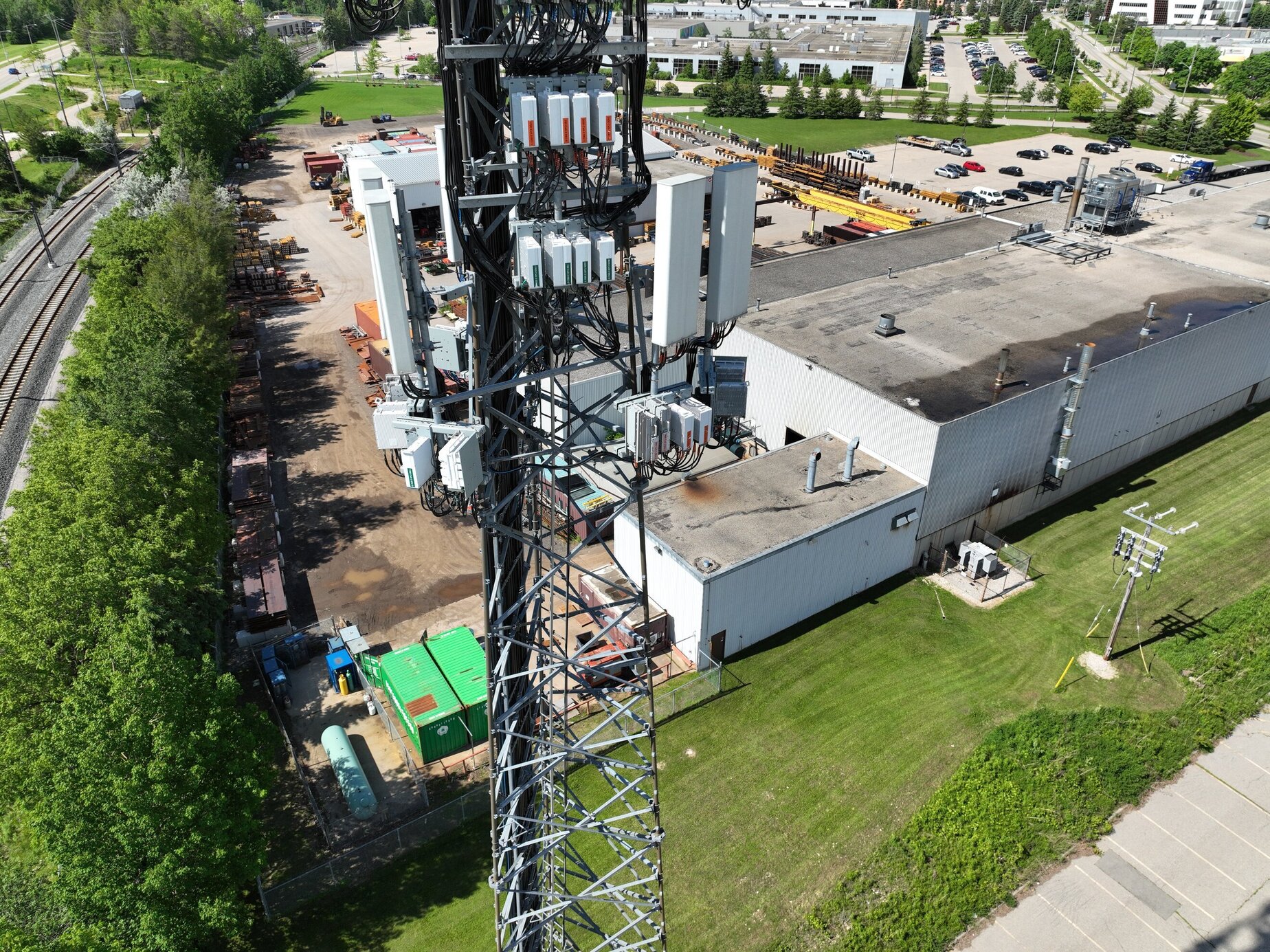} &
            \includegraphics[width=0.241\textwidth]{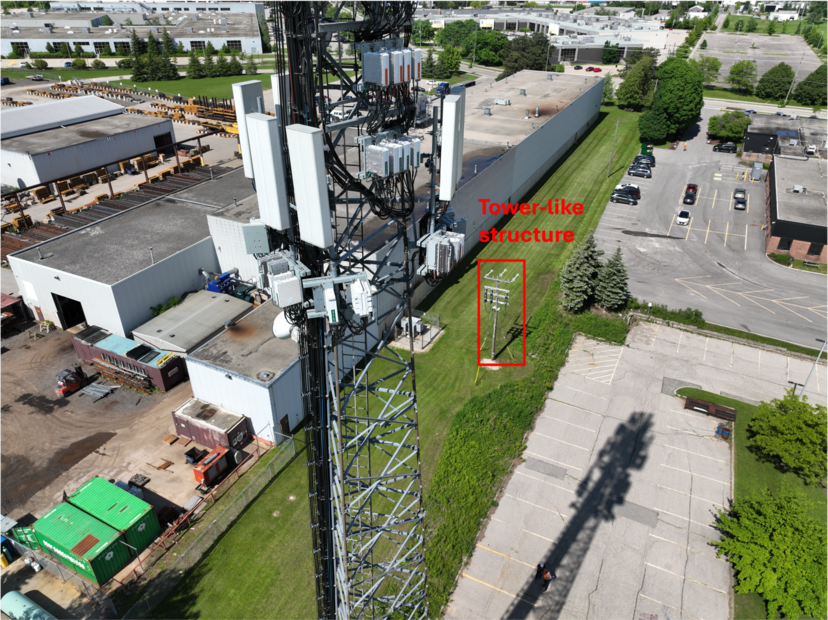} &
            \includegraphics[width=0.241\textwidth]{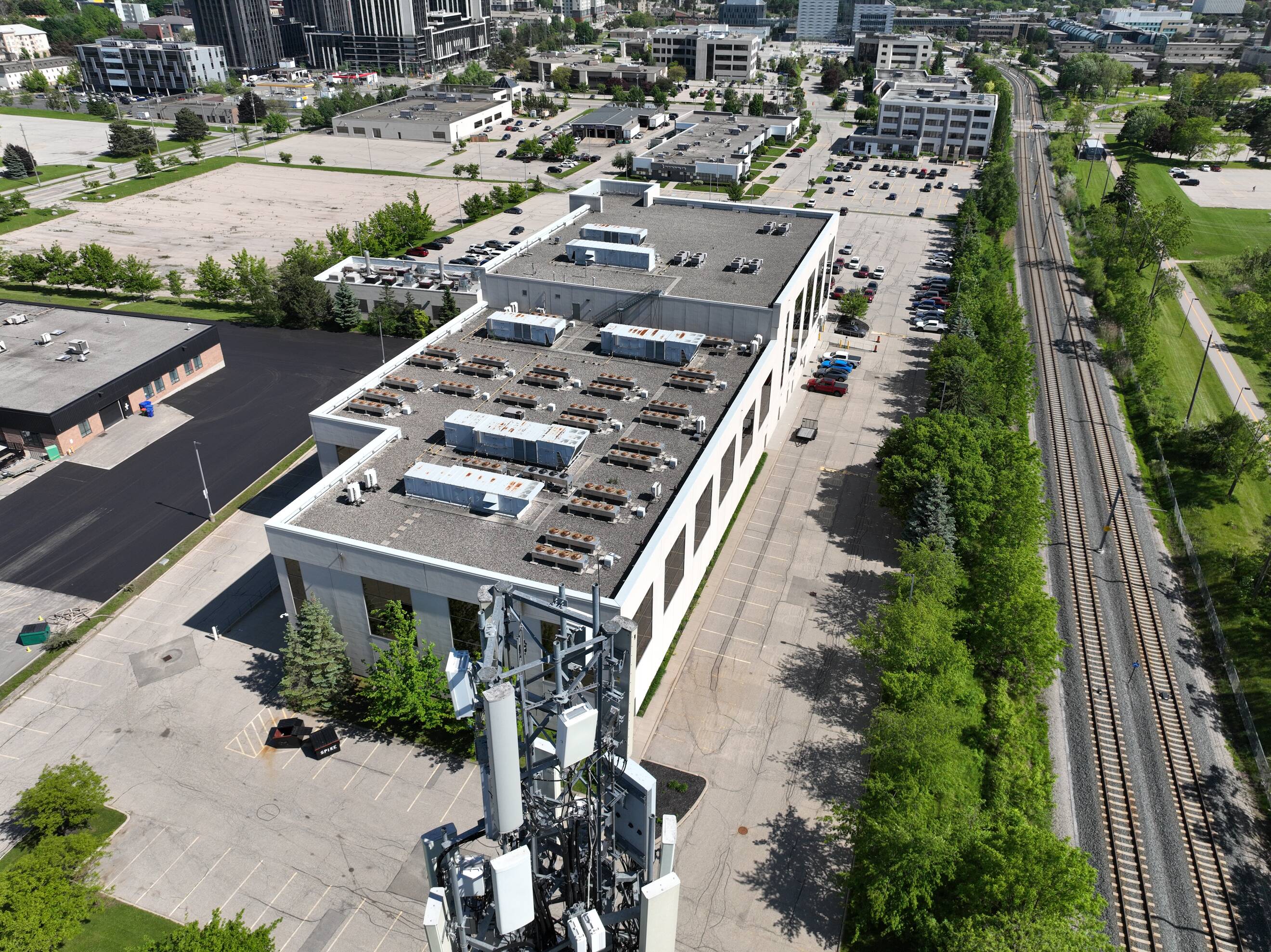} &
            \includegraphics[width=0.241\textwidth]{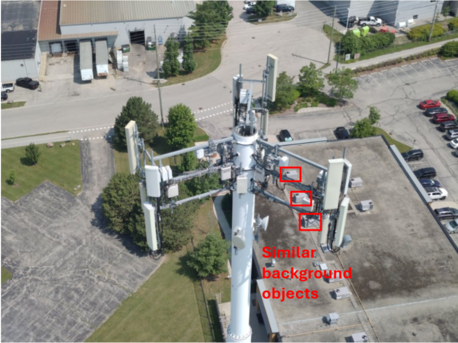} \\
            
            (a) &
            (b) &
            (c) &
            (d)
        \end{tabular}

    \caption{\textbf{Representative challenges in TOW-300.} (a) Background structures remain visible through the sparse tower lattice and exhibit colors and appearances similar to tower-mounted components. (b) A small background structure resembles the communication tower, creating ambiguity for coarse text prompts such as \textit{antenna}. (c) The tower occupies only a small portion of the full high-resolution image and is surrounded by substantial background clutter. (d) A near top-down view exposes rooftop objects through the sparse tower structure, some of which visually resemble mounted communication equipment.}
    \label{fig:dataset-challenges}
\end{figure}

\paragraph{Dataset Characteristics and Challenges.}
TOW-300 contains several characteristics, making instance segmentation challenging, as illustrated in Figure~\ref{fig:dataset-challenges}. Communication towers commonly have sparse lattice or truss structures through which vegetation, buildings, rooftops, cables, and other background objects remain visible. These background elements may exhibit shapes, textures, and colors similar to those of tower-mounted components, increasing the ambiguity between foreground components and surrounding structures. The dataset also includes substantial variation in viewpoint, scale, and perspective, including elevated and near top-down observations. In some images, the tower occupies only a small portion of the full scene, while in others, background structures resembling tower elements remain clearly visible through or around the tower. Individual components may therefore appear small, overlap with structural members, or be partially occluded by neighboring equipment, further complicating reliable instance localization and segmentation.

These challenges are particularly pronounced in a zero-shot setting. Open-vocabulary detectors rely on general-purpose textual descriptions and may not possess sufficient domain knowledge to distinguish individual antennas and radio units from the tower structure or visually similar background objects. Consequently, generic or appearance-based prompts may produce false-positive detections on rooftops, windows, signs, vegetation, and metallic structures. TOW-300 therefore provides a challenging real-world setting for evaluating zero-shot component localization and segmentation under substantial visual clutter.

\paragraph{Data Split.}
Because our proposed framework does not require model training, the dataset is divided into validation and test subsets rather than conventional training, validation, and test sets. We use 40 images to select the inference-time hyperparameters of the proposed pipeline, including the saliency--depth recovery, box-size, containment, and depth-filtering thresholds. The remaining 300 images are reserved for final evaluation. All hyperparameters are selected using only the validation subset and are subsequently fixed for evaluation on the test set.

% //////////////////////////////////////////////////////////////////
% /////////////////////////  Experiments  //////////////////////////
% //////////////////////////////////////////////////////////////////
\section{Experiments}

\subsection{Experimental Setup}

\paragraph{Compared Configurations.}
We evaluate the proposed saliency--depth conditioning strategy with two existing zero-shot segmentation frameworks: Grounded-SAM and SAM~3. For each framework, we compare the original model with its conditioned counterpart, resulting in four evaluated methods: Grounded-SAM, SD-Grounded-SAM, SAM~3, and SD-SAM~3. This paired comparison isolates the effect of saliency--depth conditioning within each framework. SD-Grounded-SAM additionally includes the proposed geometric and depth-based box-refinement stage before mask generation, whereas SD-SAM~3 relies on SAM~3's internal localization and segmentation mechanism without external box refinement.

\paragraph{Evaluation Protocol.}
The proposed methods operate in a training-free setting, and no model parameters are updated using TOW-300. All models are evaluated in inference mode using publicly available pre-trained weights. The dataset is divided into a validation subset of $N_{\mathrm{val}} = 40$ images for inference-time hyperparameter selection and a test subset of $N_{\mathrm{test}} = 300$ images for final evaluation. All hyperparameters are selected exclusively on the validation subset and fixed before evaluation on the test set. The same validation and test subsets are used for all compared methods.

\paragraph{Implementation Details.}
All experiments are conducted on the original high-resolution images without resizing, thereby preserving the fine-grained details of small tower-mounted components. Inference is performed on a single NVIDIA RTX 4090 GPU, and images are processed individually without batching. We use publicly available implementations and pre-trained checkpoints for the Transparent Background saliency model \cite{kim2022revisiting}, Depth Anything \cite{yang2024depth}, Grounding DINO \cite{liu2024grounding}, SAM \cite{kirillov2023segment}, and SAM~3 \cite{carion2025sam}.

\paragraph{Prompt and Model Configuration.}
We use the fixed textual prompt
\textit{``small bright rectangles attached to tower''}
for all compared methods. The prompt describes the visual appearance and placement of the target tower-mounted components and is kept unchanged across methods to avoid introducing differences caused by prompt selection.

For Grounded-SAM and SD-Grounded-SAM, Grounding DINO with a Swin-T backbone is used for open-vocabulary detection, followed by SAM ViT-H for mask generation, whereas SAM~3 and SD-SAM~3 are evaluated through SAM~3's zero-shot concept-prompted segmentation pipeline. The conditioned variants share the same foreground conditioning configuration, using  Depth Anything ViT-L for monocular depth estimation and the base Transparent Background model for saliency prediction.

\paragraph{Hyperparameter Selection.}
The saliency--depth conditioning module and the SD-Grounded-SAM box-refinement stage involve several inference-time hyperparameters introduced in the Method section. These include the depth-recovery threshold $\tau_{\mathrm{rec}}$, the relative box-size threshold $\tau_s$, the containment threshold $\tau_{\mathrm{iou}}$, and the box-level depth threshold $\tau_d$. The values selected on the validation subset are
\begin{equation}
\tau_{\mathrm{rec}} = 140,\qquad
\tau_s = 0.4,\qquad
\tau_{\mathrm{iou}} = 0.5,\qquad
\tau_d = 70.
\end{equation}
These values are selected using only the validation subset and remain fixed during final testing. Each segmentation framework also uses a model-specific confidence threshold to determine which predicted instances are retained. We select this threshold through a grid search on the validation subset, using the value that maximizes the F1-score at an instance-mask IoU threshold of $0.5$. The same selection procedure is applied to the original and conditioned variants of each framework. The resulting confidence thresholds are $0.14$ for Grounded-SAM and SD-Grounded-SAM, and $0.20$ for SAM~3 and SD-SAM~3. These thresholds are also fixed during final evaluation.

\subsection{Evaluation Metrics}

We evaluate performance using both instance-level and semantic segmentation metrics to provide a comprehensive assessment of detection and mask quality.

\paragraph{Instance Segmentation Metrics.}
We report standard instance segmentation metrics, including Average Precision (AP) at different IoU thresholds. Specifically, we use AP@0.5, AP@0.75, and the COCO-style mean Average Precision (mAP@[0.5:0.95]) \cite{lin2014microsoft}. These metrics evaluate the quality of both localization and segmentation across varying levels of overlap between predicted and ground-truth instances. In addition, we report Precision@0.5 and Recall@0.5 to analyze detection performance at IoU=0.5. To further assess mask quality for correctly matched instances, we compute the mean matched IoU, which measures the average overlap between predicted and ground-truth masks for true positive detections.

\paragraph{Semantic Segmentation Metrics.}
To evaluate overall mask quality at the image level, we also compute semantic segmentation metrics by merging all predicted instance masks into a single binary mask. We report IoU and Dice coefficient for the \textit{tower-mounted components} class. These metrics capture the overall coverage and accuracy of the predicted foreground regions.

\subsection{Quantitative Results}
\begin{table}[t]
  \centering
  \caption{Instance segmentation performance on the TOW-300 test set. Mean matched IoU is computed over true-positive instance matches at an IoU threshold of 0.5. Best results are shown in bold, and second-best results are underlined.}
  \resizebox{\linewidth}{!}{
    \begin{tabular}{lcccccc}
    \toprule
    \multirow{2}[2]{*}{Method}  
    & \multirow{2}[2]{*}{mAP@[0.5:0.95] ($\uparrow$)} 
    & \multirow{2}[2]{*}{AP@0.75 ($\uparrow$)}    
    & \multicolumn{4}{c}{@IoU=0.5} \\
    
    \cmidrule{4-7}
    & & 
    & AP ($\uparrow$)  
    & Recall ($\uparrow$)  
    & Precision ($\uparrow$)  
    & Matched IoU ($\uparrow$) \\
    
    \midrule
    Grounded-SAM      
    & 0.1737
    & 0.1819
    & 0.2236                
    & 0.3971                    
    & 0.4962                       
    & 0.8879 \\
    
    SAM~3             
    & \underline{0.4250}
    & \underline{0.4831}
    & \underline{0.5438}                
    & \underline{0.6089}                    
    & 0.7662                
    & 0.8787 \\
    
    \midrule
    
    SD-Grounded-SAM (Ours)
    & 0.3780
    & 0.4129
    & 0.4751                
    & 0.5073
    & \textbf{0.8970}                       
    & \underline{0.8989} \\

    SD-SAM~3 (Ours)
    & \textbf{0.5701}
    & \textbf{0.6270}
    & \textbf{0.6709}                
    & \textbf{0.7311}                    
    & \underline{0.7817}                       
    & \textbf{0.9107} \\
    
    \bottomrule
    \end{tabular}
  }
  \label{tab:instance-results}
\end{table}

\begin{table}[t]
  \centering
  \caption{Semantic segmentation results obtained by merging all predicted instance masks into a binary foreground mask. Best results are shown in bold, and second-best results are underlined.}
  \resizebox{0.45\linewidth}{!}{
  \begin{tabular}{lcc}
    \toprule
    Method & IoU ($\uparrow$) & Dice ($\uparrow$) \\
    \midrule     
    Grounded-SAM      
    & 0.2876
    & 0.4467 \\
    
    SAM~3             
    & 0.3630
    & 0.5326 \\
    
    \midrule     
    
    SD-Grounded-SAM (Ours)   
    & \textbf{0.6281}
    & \textbf{0.7716} \\

    SD-SAM~3 (Ours)
    & \underline{0.4376}
    & \underline{0.6088} \\
    
    \bottomrule
  \end{tabular}
  }
  \label{tab:semantic-results}
\end{table}

Tables~\ref{tab:instance-results} and~\ref{tab:semantic-results} report the instance-level and image-level segmentation results, respectively. The proposed saliency--depth conditioning strategy improves both evaluated zero-shot segmentation frameworks across all reported metrics. SD-SAM~3 achieves the strongest overall instance segmentation performance, obtaining the highest mAP, AP@0.75, AP@0.5, recall, and mean matched IoU. In contrast, SD-Grounded-SAM achieves the highest precision and the strongest image-level semantic segmentation results. These findings show that foreground conditioning provides consistent benefits across architecturally distinct segmentation systems, while the downstream framework and box-refinement design influence the resulting precision--recall characteristics.

\paragraph{Effect on Grounded-SAM.}
SD-Grounded-SAM substantially improves its direct Grounded-SAM baseline across every instance-level metric. COCO-style mAP increases from 0.1737 to 0.3780, while AP@0.5 increases from 0.2236 to 0.4751 and AP@0.75 increases from 0.1819 to 0.4129. These correspond to relative improvements of approximately 117.6\%, 112.5\%, and 127.0\%, respectively. Recall also increases from 0.3971 to 0.5073, indicating that the conditioned pipeline recovers more valid component instances than the original Grounded-SAM configuration.

The largest improvement is observed in precision, which increases from 0.4962 to 0.8970. This result indicates that the combined foreground-conditioning and box-refinement stages effectively suppress false detections arising from visually similar regions. Mean matched IoU also increases from 0.8879 to 0.8989. Although this gain is smaller, it shows that the refined box prompts preserve, and slightly improve, the quality of masks associated with correctly localized instances. Overall, the results suggest that the main weakness of Grounded-SAM in this domain lies in candidate localization and prompt generation rather than in the mask-generation capability of SAM itself.

\paragraph{Effect on SAM~3.}
Applying the same saliency--depth conditioning strategy to SAM~3 also produces consistent improvements, despite the model performing localization and segmentation internally and receiving no external box refinement. SD-SAM~3 increases mAP from 0.4250 to 0.5701, AP@0.5 from 0.5438 to 0.6709, and AP@0.75 from 0.4831 to 0.6270. These correspond to relative gains of approximately 34.1\%, 23.4\%, and 29.8\%, respectively. Recall increases from 0.6089 to 0.7311, showing that conditioning allows SAM~3 to identify a substantially larger proportion of the annotated components.

Precision also improves from 0.7662 to 0.7817, while mean matched IoU increases from 0.8787 to 0.9107. The simultaneous gains in precision and recall indicate that foreground conditioning improves the quality of concept localization rather than simply making the model more conservative or more permissive. By suppressing irrelevant visual content while preserving tower-relevant regions, the conditioned input helps SAM~3 identify more valid components and generate more accurate masks. Since SD-SAM~3 does not use the external box-refinement stage of SD-Grounded-SAM, these improvements can be attributed directly to the saliency--depth conditioning module.

\paragraph{Comparison of the Conditioned Frameworks.}
Among the conditioned methods, SD-SAM~3 achieves higher instance-level AP and recall than SD-Grounded-SAM. This indicates that SAM~3's integrated concept-localization mechanism is more effective at discovering small, partially occluded, or visually weak component instances once background interference has been reduced.

SD-Grounded-SAM, however, achieves the highest precision. This difference is consistent with its explicit box-refinement stage, which removes detections based on tower-relative size, pairwise containment, and relative depth before mask generation. Consequently, SD-Grounded-SAM adopts a more conservative operating profile, retaining fewer predictions but producing a larger proportion of valid component detections. The two conditioned configurations are therefore complementary: SD-SAM~3 is preferable when component discovery and recall are prioritized, whereas SD-Grounded-SAM is advantageous when false-positive suppression is more important.

\paragraph{Image-Level Foreground Quality.}
The semantic results in Table~\ref{tab:semantic-results} further demonstrate the effect of the proposed conditioning strategy. SD-Grounded-SAM improves semantic IoU from 0.2876 to 0.6281 and Dice from 0.4467 to 0.7716 relative to Grounded-SAM. These correspond to relative improvements of approximately 118.4\% and 72.7\%, respectively. The large gains indicate that foreground conditioning and explicit box refinement substantially reduce the total area assigned to irrelevant background objects.

SD-SAM~3 also improves consistently over SAM~3, increasing semantic IoU from 0.3630 to 0.4376 and Dice from 0.5326 to 0.6088. These gains of approximately 20.6\% and 14.3\% show that conditioning produces cleaner aggregate masks even when object localization remains internal to SAM~3. However, SD-Grounded-SAM achieves the strongest semantic results overall.

This difference between the instance-level and semantic results can be explained by the different prediction behavior of the two pipelines. SD-Grounded-SAM is deliberately conservative: its box-refinement stage removes many uncertain detections, including oversized, nested, and distant boxes, and therefore retains a smaller set of high-confidence component predictions. As a result, it may miss some valid instances, but it also produces substantially fewer false-positive masks, which leads to cleaner merged foreground predictions and stronger semantic IoU and Dice. In contrast, SD-SAM~3 attempts to recover a larger number of component instances and therefore achieves stronger instance-level performance, but this higher recall can also introduce additional false-positive or overly broad masks. In some cases, SD-SAM~3 predicts a large mask covering much of the tower in addition to the desired component-level masks, which increases the predicted foreground area and degrades image-level semantic metrics. A representative example of this behavior is shown in Fig.~\ref{fig:qualitative-results}.

\paragraph{Discussion.}
Taken together, the results support two main conclusions. First, saliency--depth foreground conditioning addresses a general background-ambiguity problem rather than a limitation specific to a single segmentation architecture. Both Grounded-SAM and SAM~3 improve across instance- and image-level metrics when supplied with the conditioned image. Second, the downstream segmentation framework determines how these benefits are expressed. SD-SAM~3 provides the strongest overall instance discovery and mask overlap, while SD-Grounded-SAM combines conditioning with explicit box refinement to achieve the highest precision and cleanest aggregate foreground segmentation. These complementary behaviors provide different operating points for practical inspection workflows, depending on whether maximizing component coverage or minimizing false detections is the primary objective.

\subsection{Qualitative Results}

\begin{figure}[t]
    \centering
    \setlength{\tabcolsep}{2pt}
    \renewcommand{\arraystretch}{1.05}
    \begin{tabular}{
        C{0.158\textwidth}
        C{0.158\textwidth}
        C{0.158\textwidth}
        C{0.158\textwidth}
        C{0.158\textwidth}
        C{0.158\textwidth}
    }
        \small Input Image &
        \small Ground Truth &
        \small Grounded-SAM &
        \small SAM~3 &
        \small SD-Grounded-SAM (Ours) &
        \small SD-SAM~3 (Ours) \\

        \includegraphics[width=0.158\textwidth]{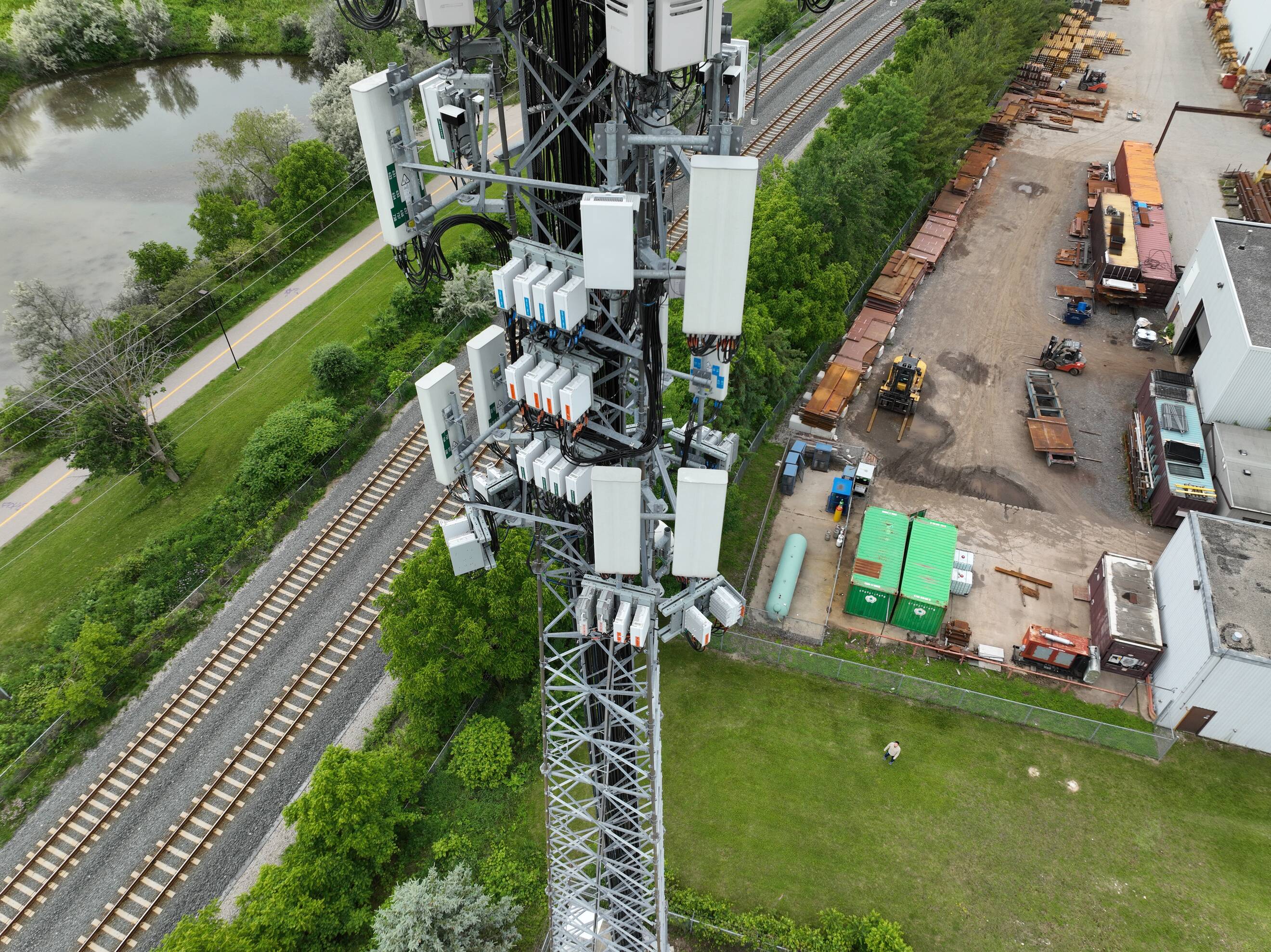} &
        \includegraphics[width=0.158\textwidth]{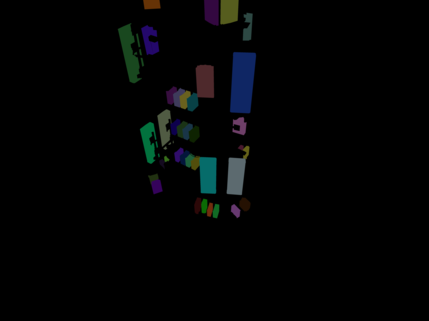} &
        \includegraphics[width=0.158\textwidth]{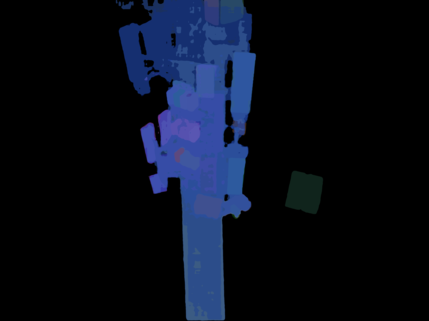} &
        \includegraphics[width=0.158\textwidth]{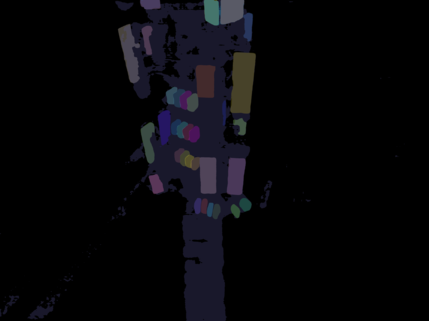} &
        \includegraphics[width=0.158\textwidth]{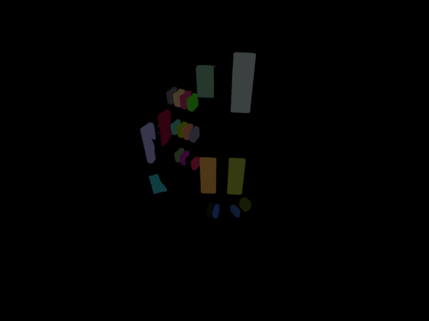} &
        \includegraphics[width=0.158\textwidth]{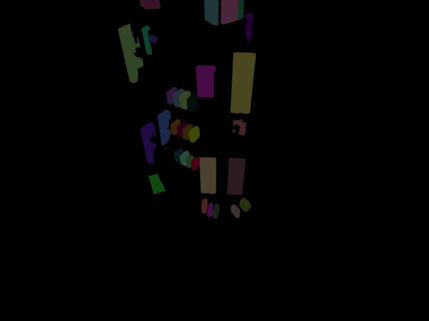} \\

        \includegraphics[width=0.158\textwidth]{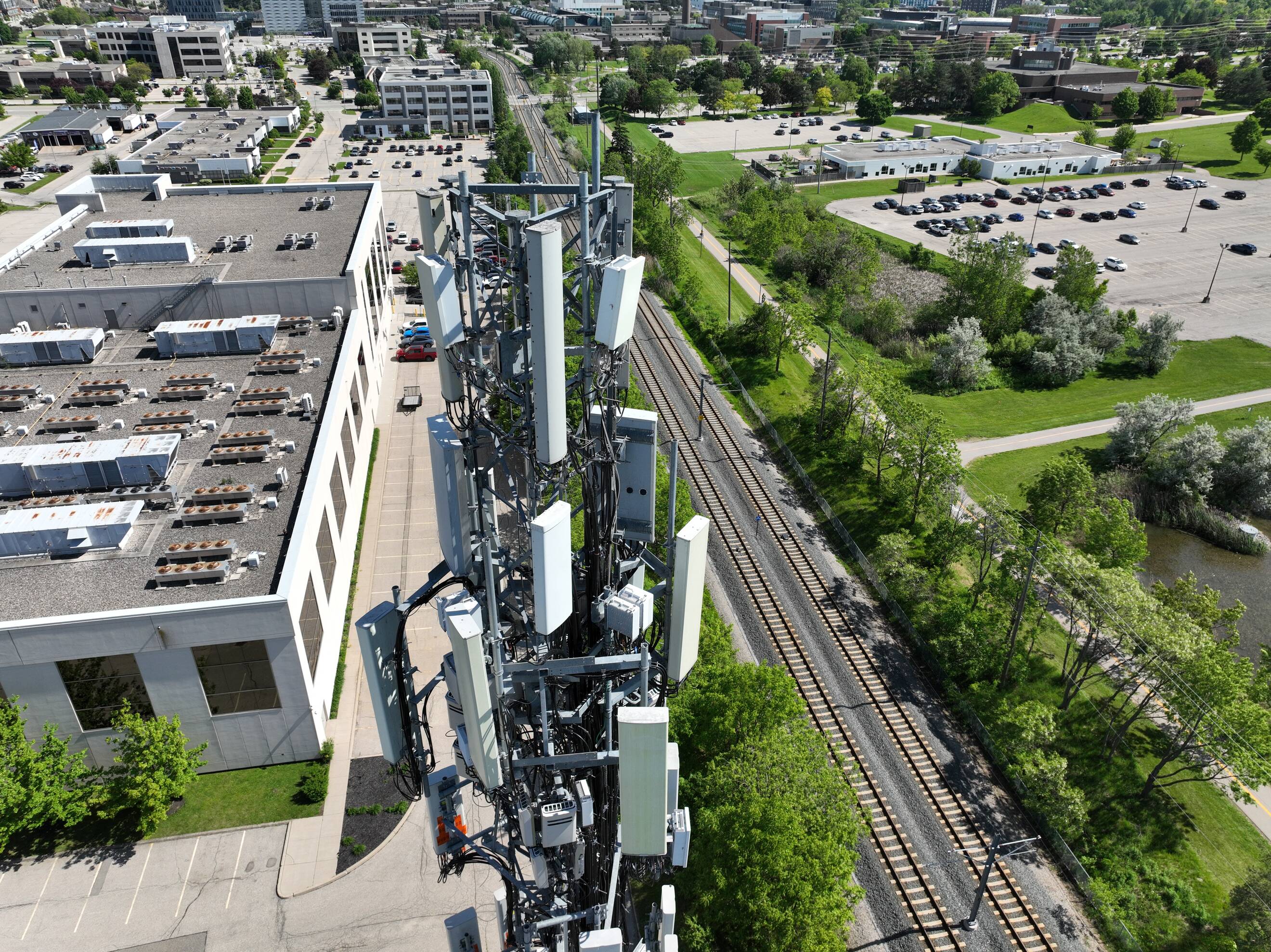} &
        \includegraphics[width=0.158\textwidth]{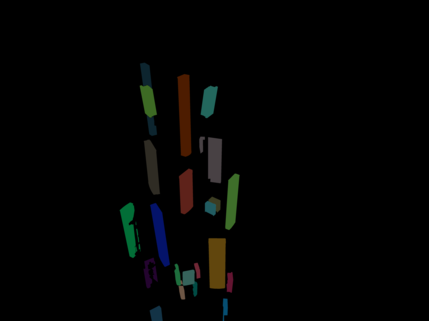} &
        \includegraphics[width=0.158\textwidth]{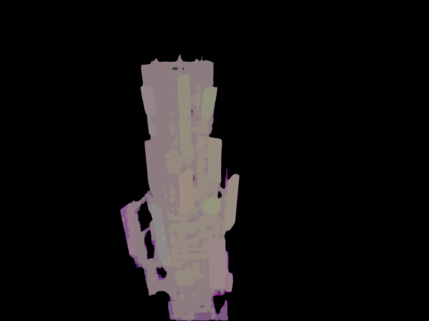} &
        \includegraphics[width=0.158\textwidth]{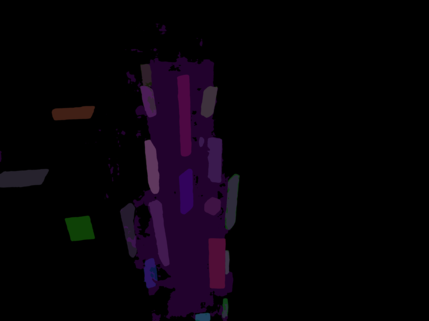} &
        \includegraphics[width=0.158\textwidth]{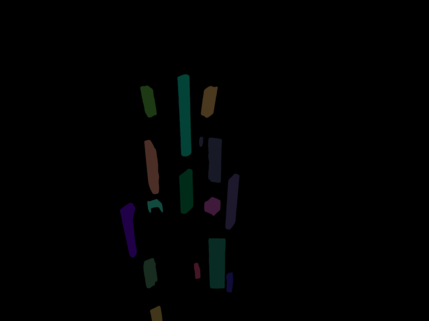} &
        \includegraphics[width=0.158\textwidth]{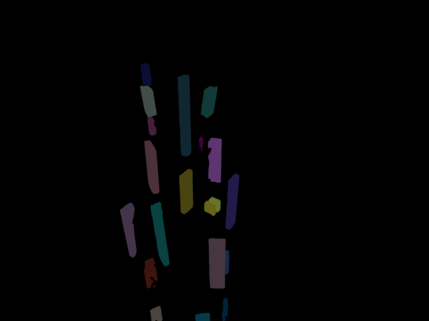} \\

        \includegraphics[width=0.158\textwidth]{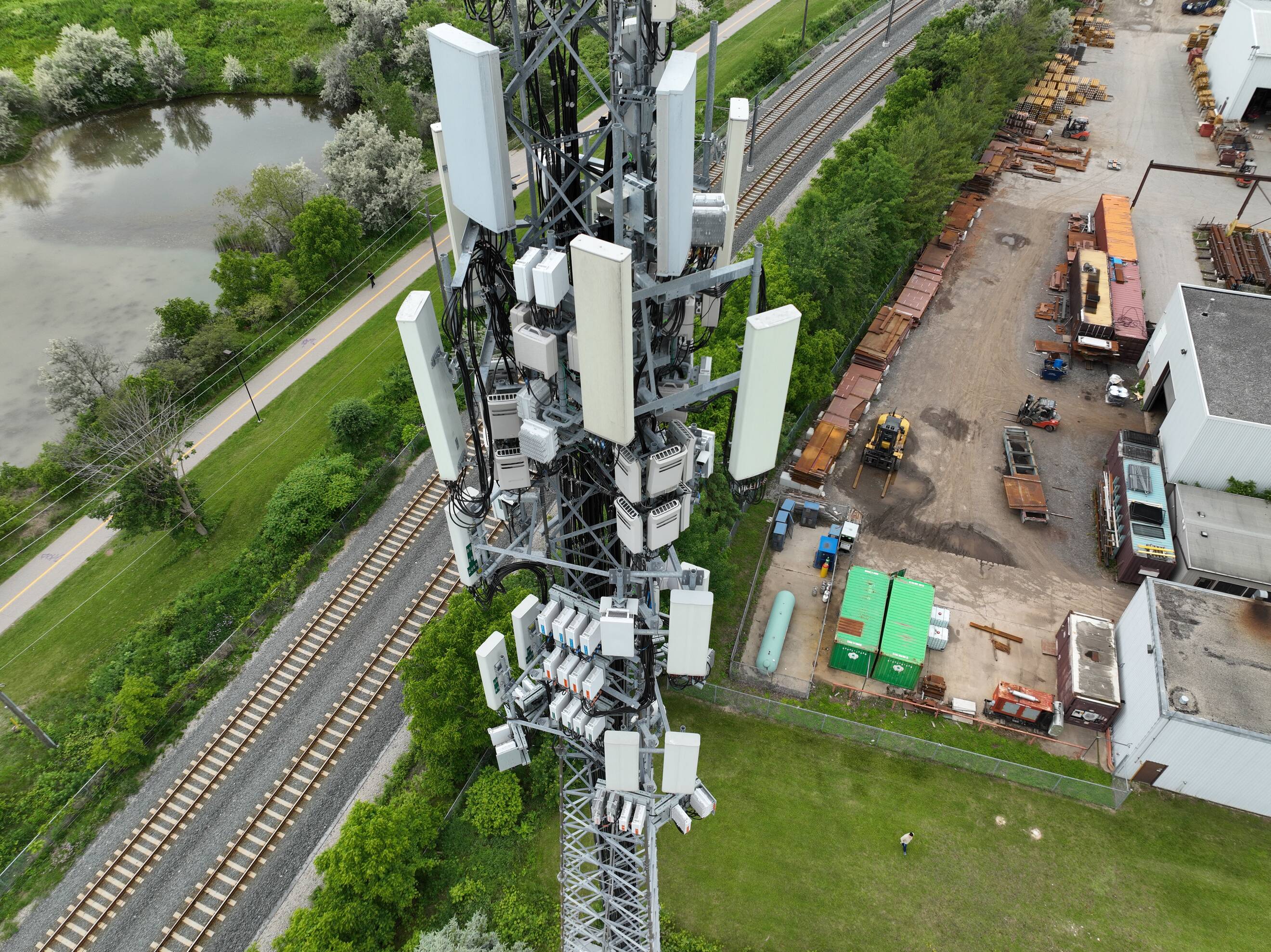} &
        \includegraphics[width=0.158\textwidth]{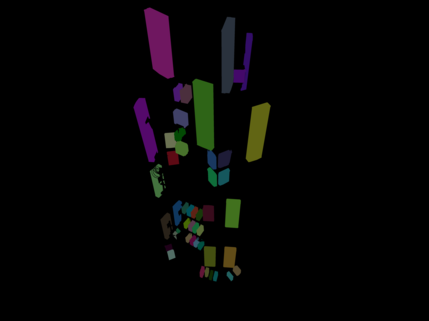} &
        \includegraphics[width=0.158\textwidth]{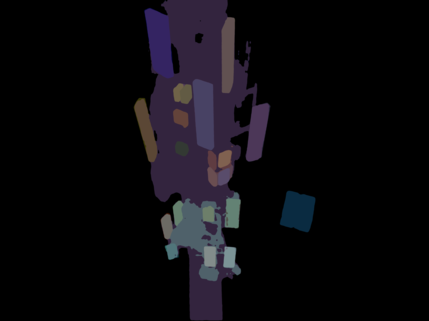} &
        \includegraphics[width=0.158\textwidth]{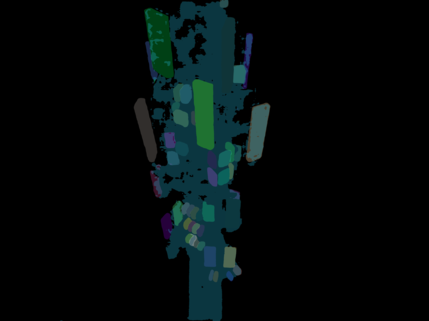} &
        \includegraphics[width=0.158\textwidth]{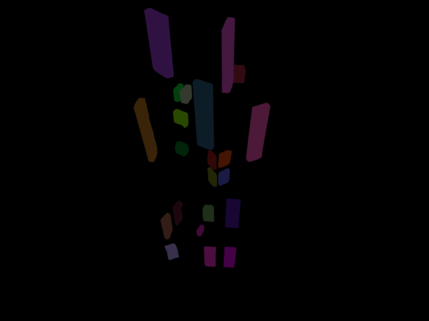} &
        \includegraphics[width=0.158\textwidth]{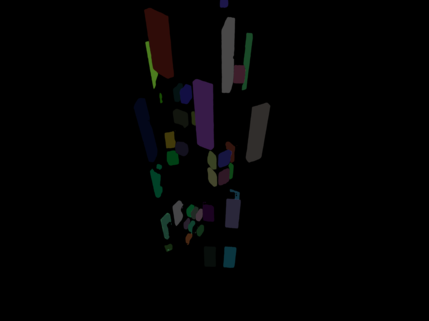} \\

        \includegraphics[width=0.158\textwidth]{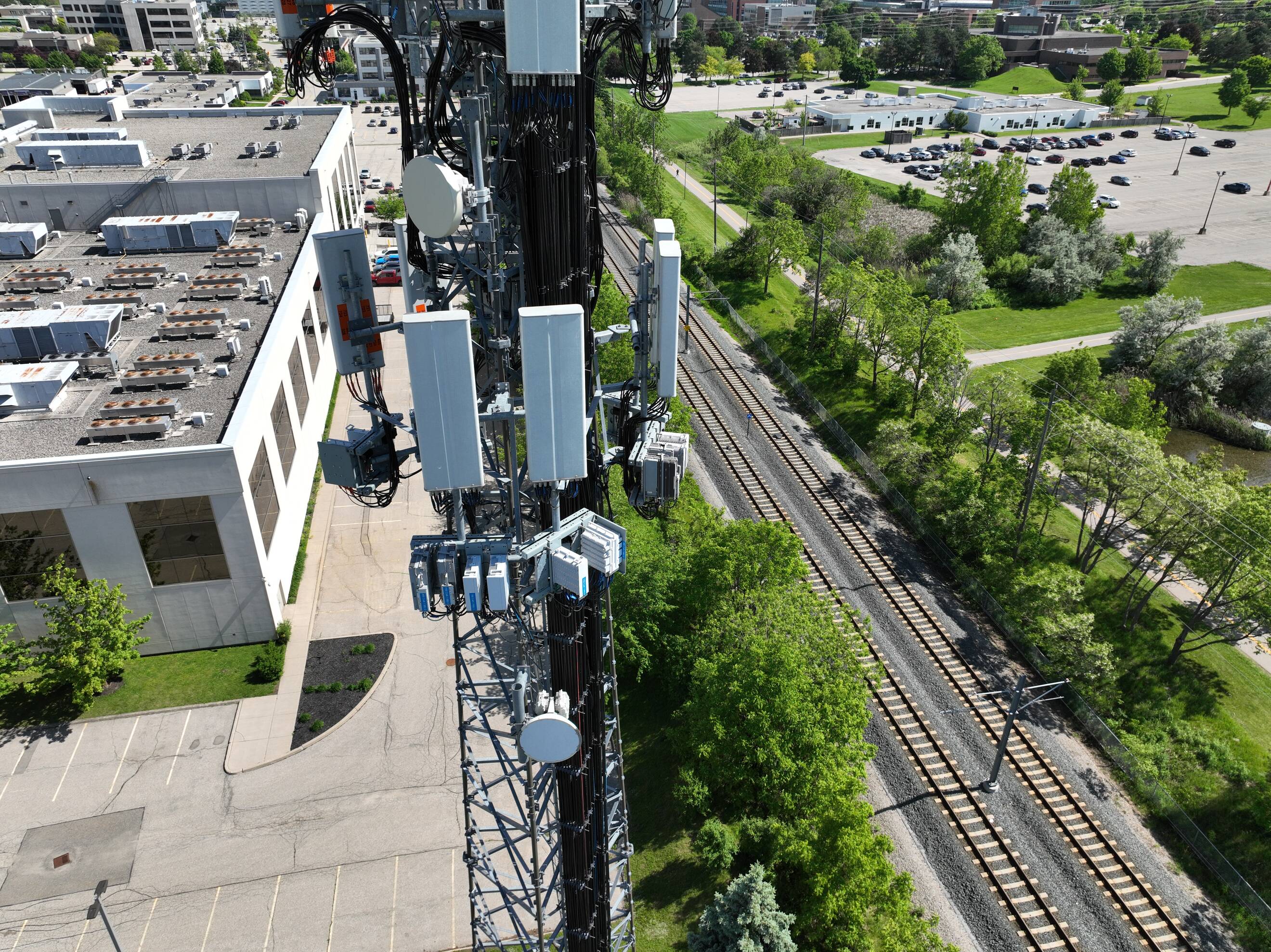} &
        \includegraphics[width=0.158\textwidth]{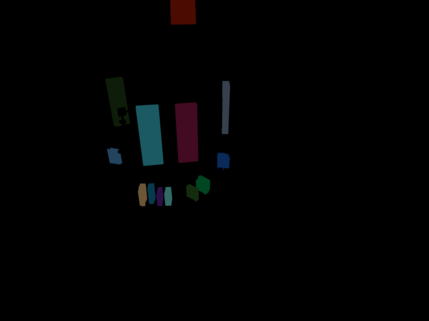} &
        \includegraphics[width=0.158\textwidth]{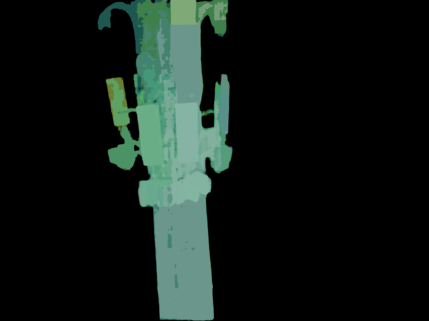} &
        \includegraphics[width=0.158\textwidth]{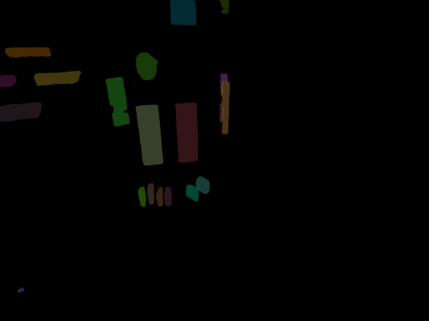} &
        \includegraphics[width=0.158\textwidth]{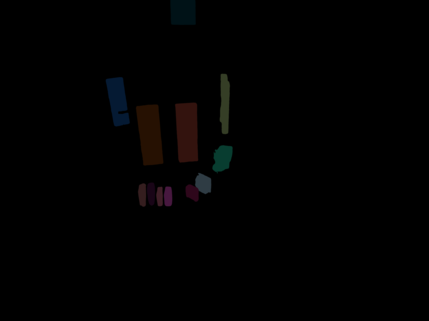} &
        \includegraphics[width=0.158\textwidth]{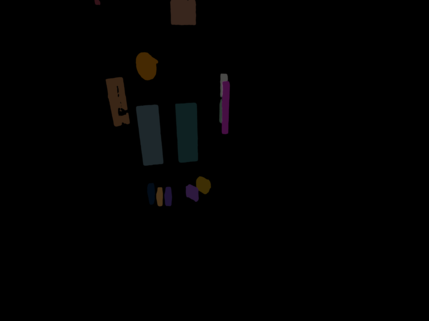} \\

        \includegraphics[width=0.158\textwidth]{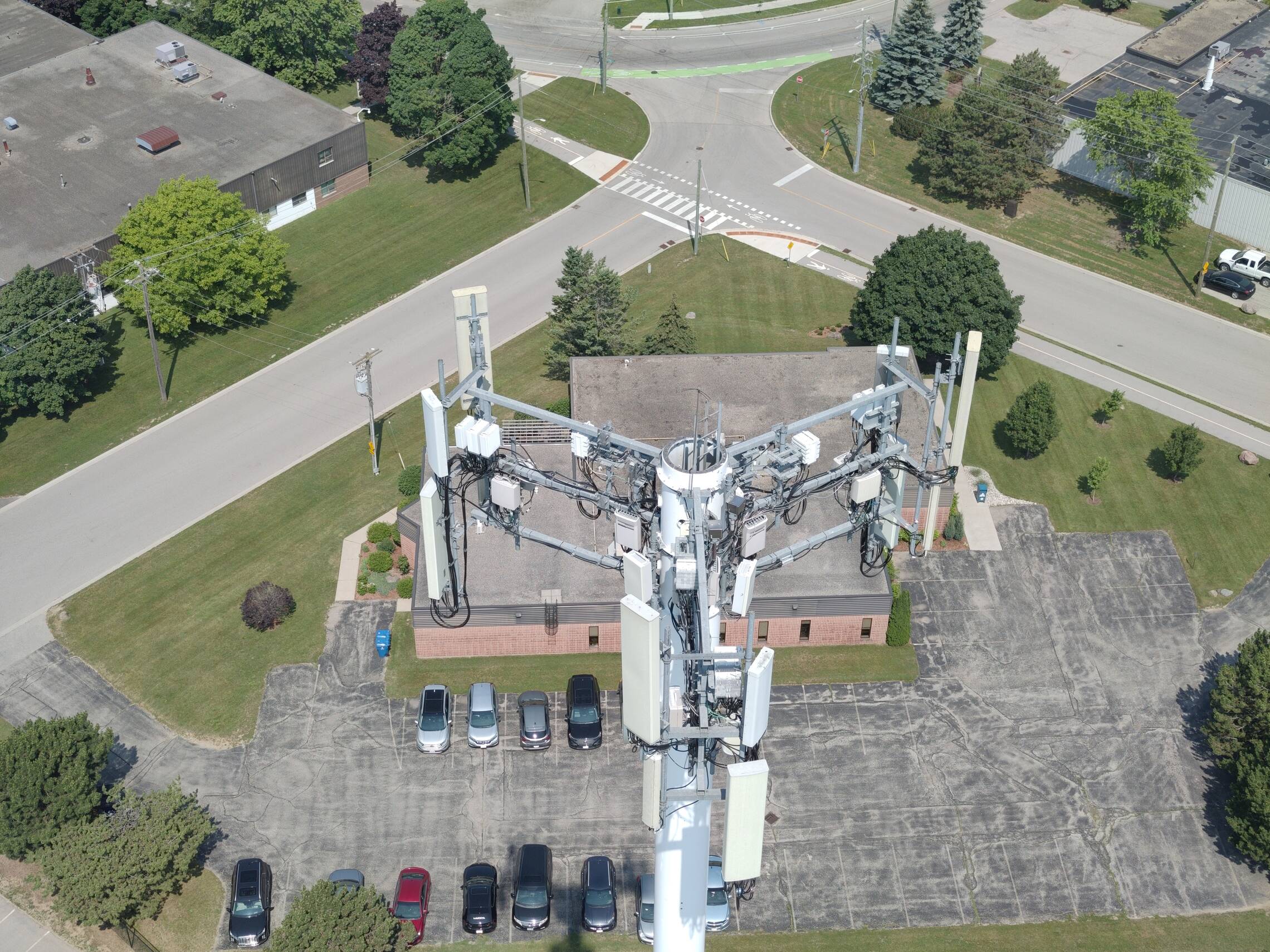} &
        \includegraphics[width=0.158\textwidth]{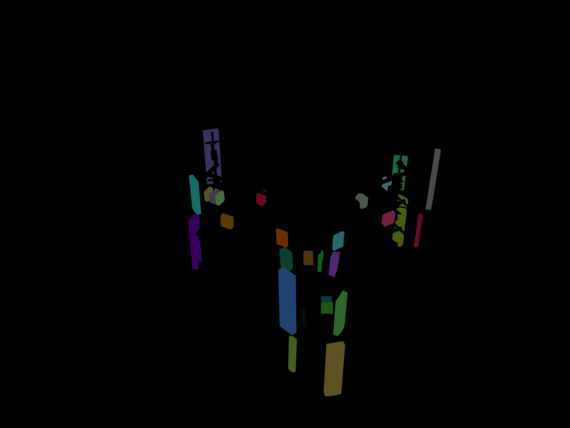} &
        \includegraphics[width=0.158\textwidth]{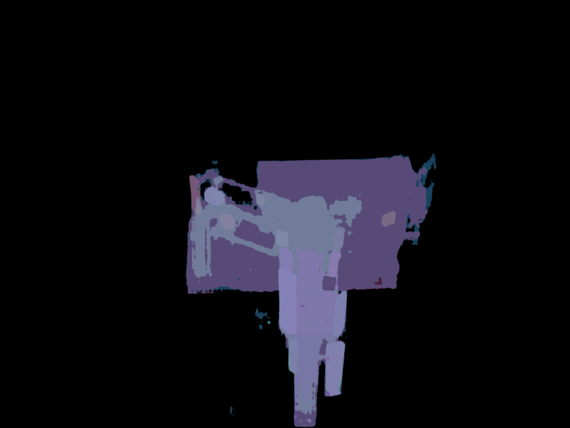} &
        \includegraphics[width=0.158\textwidth]{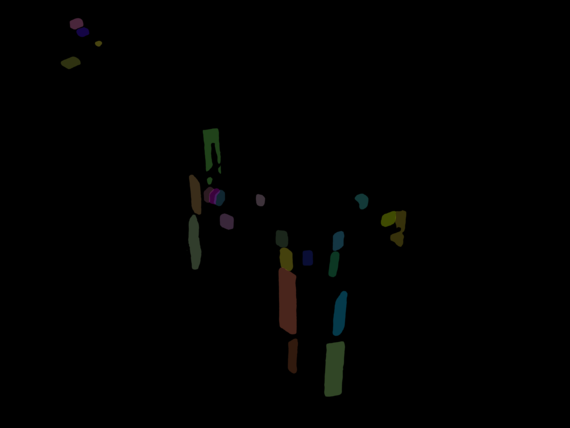} &
        \includegraphics[width=0.158\textwidth]{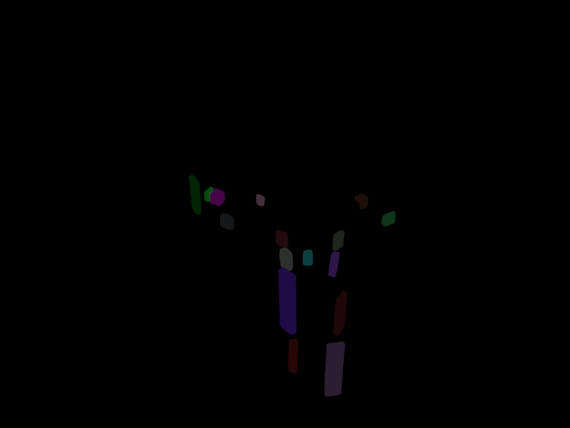} &
        \includegraphics[width=0.158\textwidth]{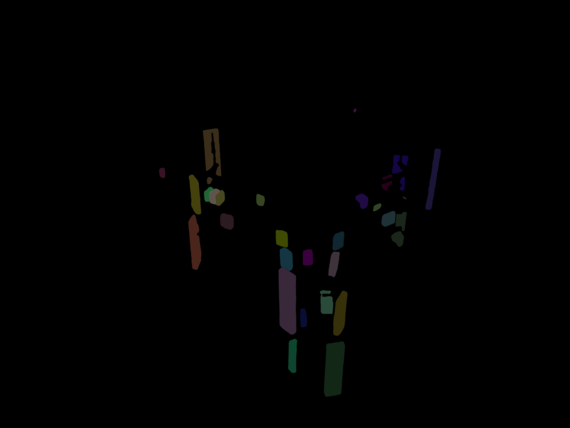} \\

        \includegraphics[width=0.158\textwidth]{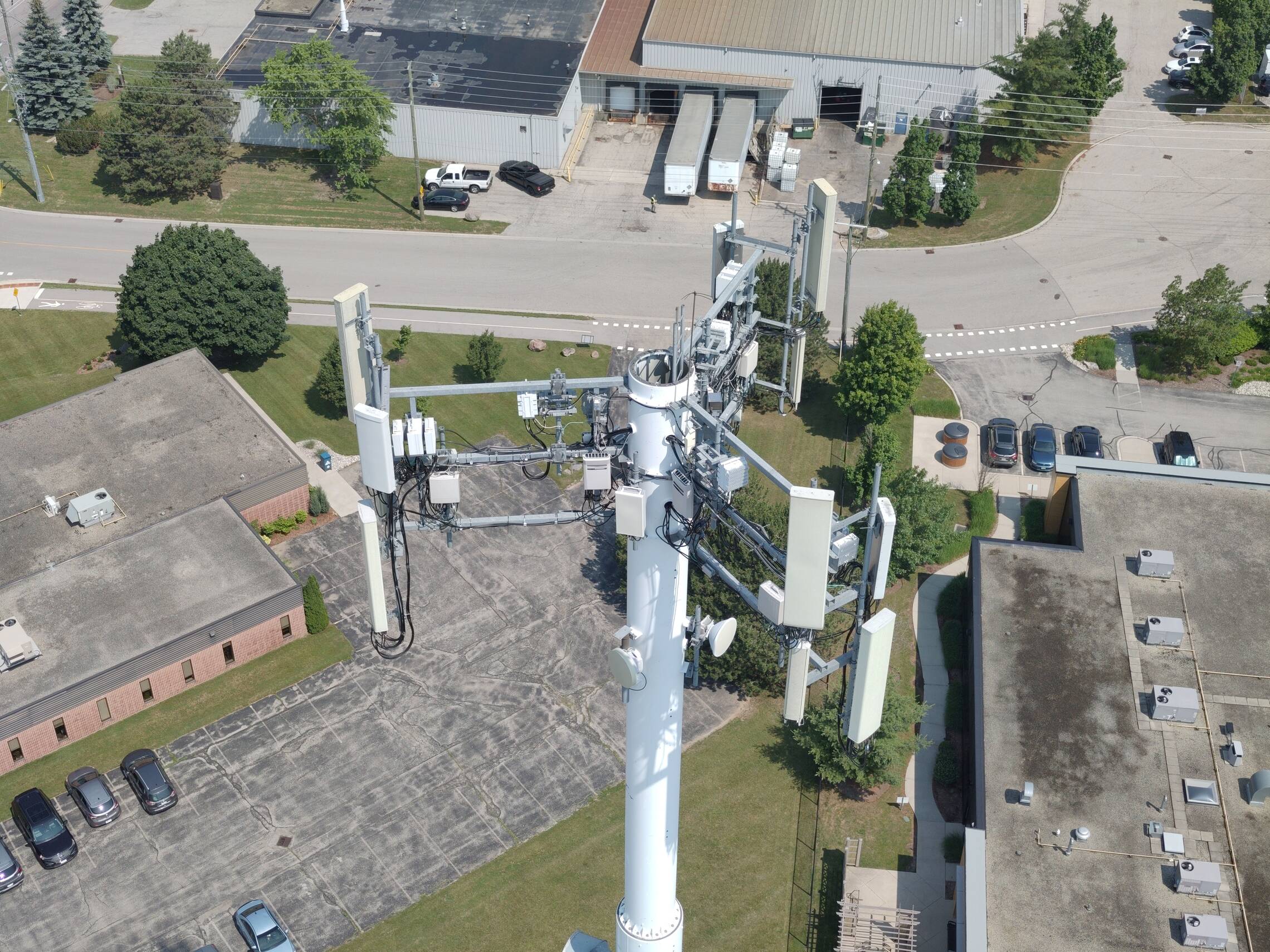} &
        \includegraphics[width=0.158\textwidth]{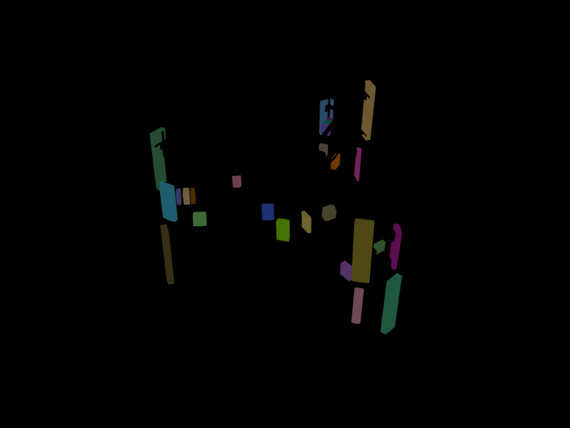} &
        \includegraphics[width=0.158\textwidth]{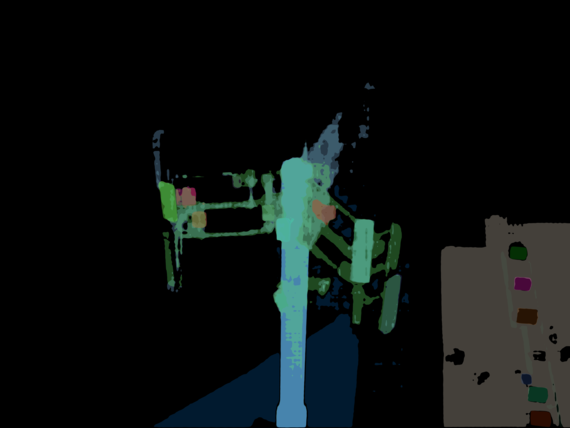} &
        \includegraphics[width=0.158\textwidth]{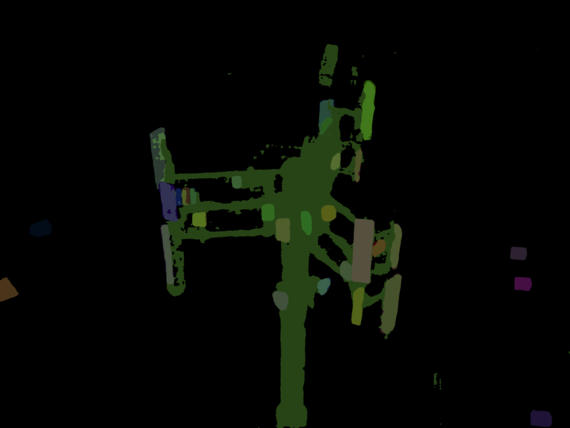} &
        \includegraphics[width=0.158\textwidth]{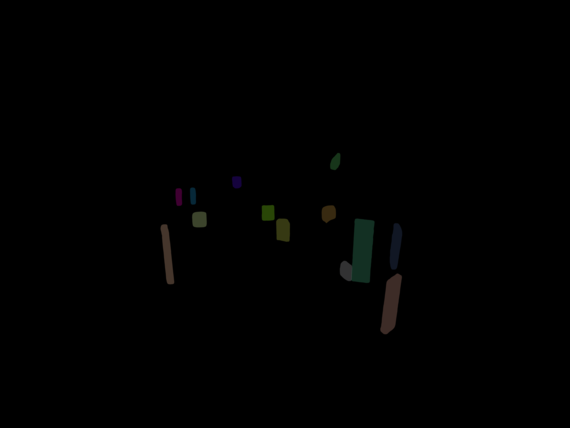} &
        \includegraphics[width=0.158\textwidth]{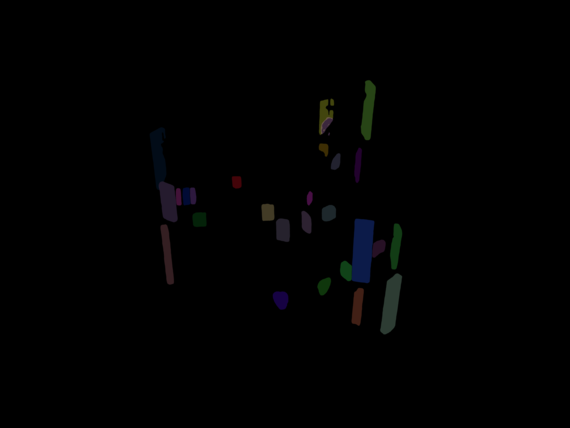}

    \end{tabular}

    \caption{Qualitative comparison of Grounded-SAM, SAM~3, and their saliency--depth-conditioned variants. Each row presents a representative UAV image, its ground-truth annotation, and the corresponding predictions.}
    \label{fig:qualitative-results}
\end{figure}

\begin{figure}[h!]
    \centering
    \setlength{\tabcolsep}{2pt}
    \renewcommand{\arraystretch}{1.05}
    \begin{tabular}{
        C{0.158\textwidth}
        C{0.158\textwidth}
        C{0.158\textwidth}
        C{0.158\textwidth}
        C{0.158\textwidth}
        C{0.158\textwidth}
    }
        \small Input Image &
        \small Ground Truth &
        \small Grounded-SAM &
        \small SAM~3 &
        \small SD-Grounded-SAM (Ours) &
        \small SD-SAM~3 (Ours) \\
        
        \includegraphics[width=0.158\textwidth]{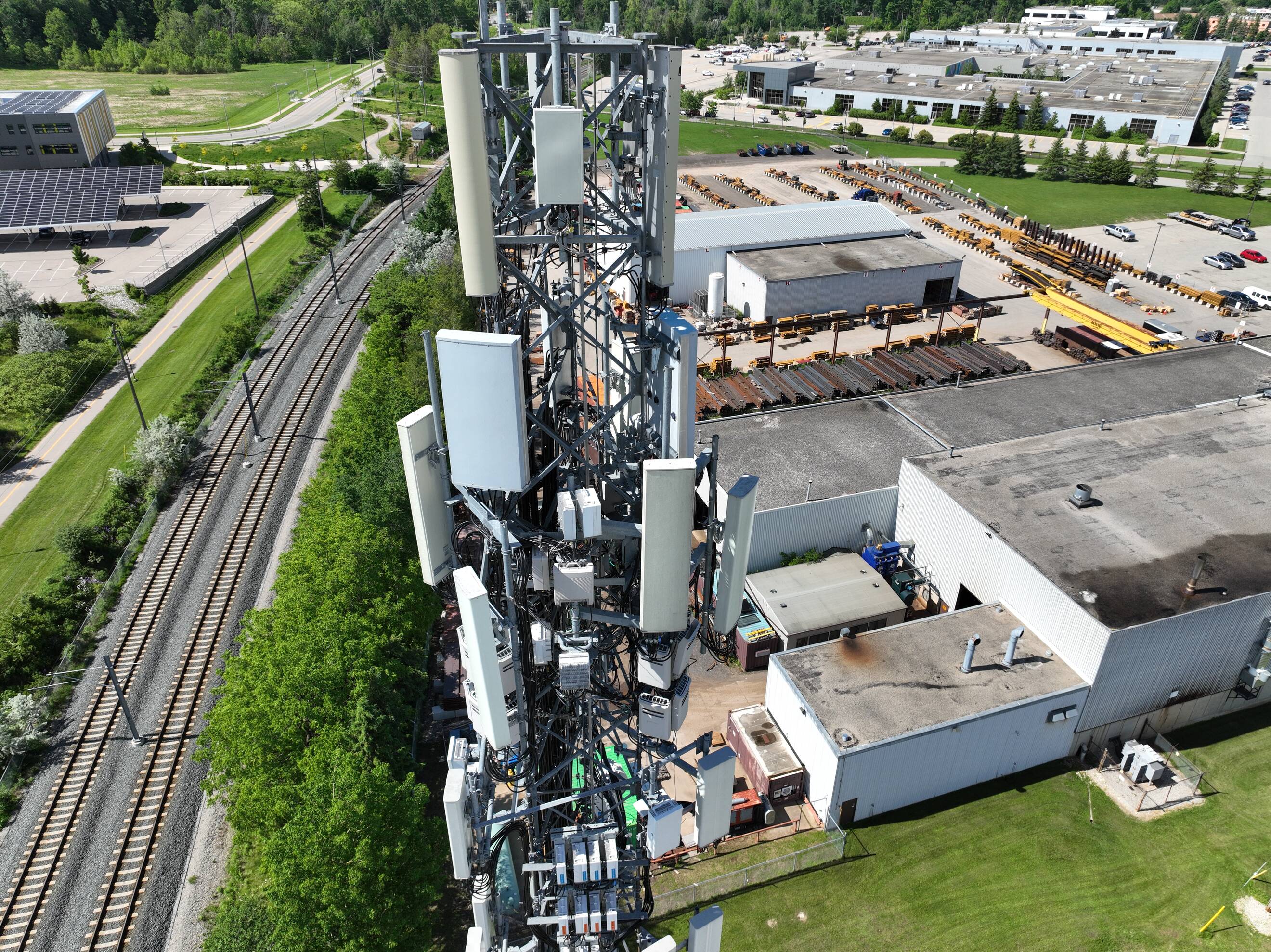} &
        \includegraphics[width=0.158\textwidth]{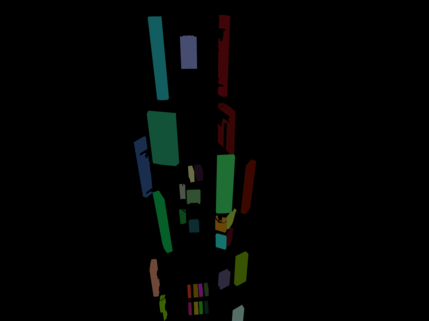} &
        \includegraphics[width=0.158\textwidth]{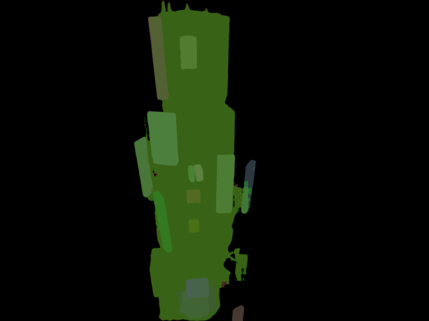} &
        \includegraphics[width=0.158\textwidth]{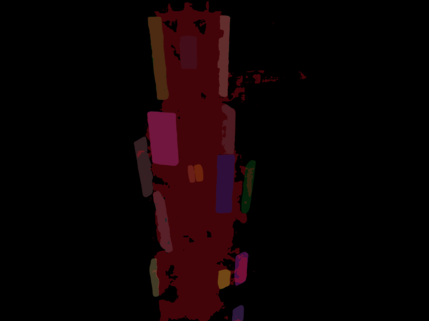} &
        \includegraphics[width=0.158\textwidth]{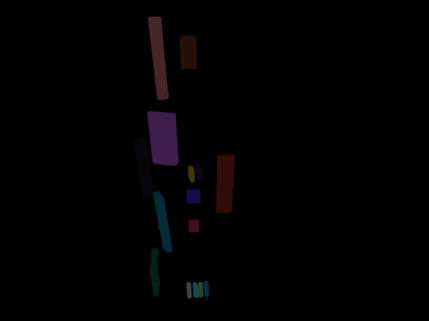} &
        \includegraphics[width=0.158\textwidth]{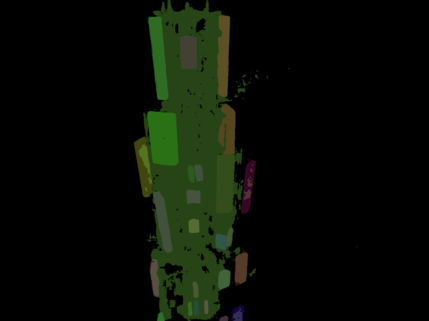} \\

        \includegraphics[width=0.158\textwidth]{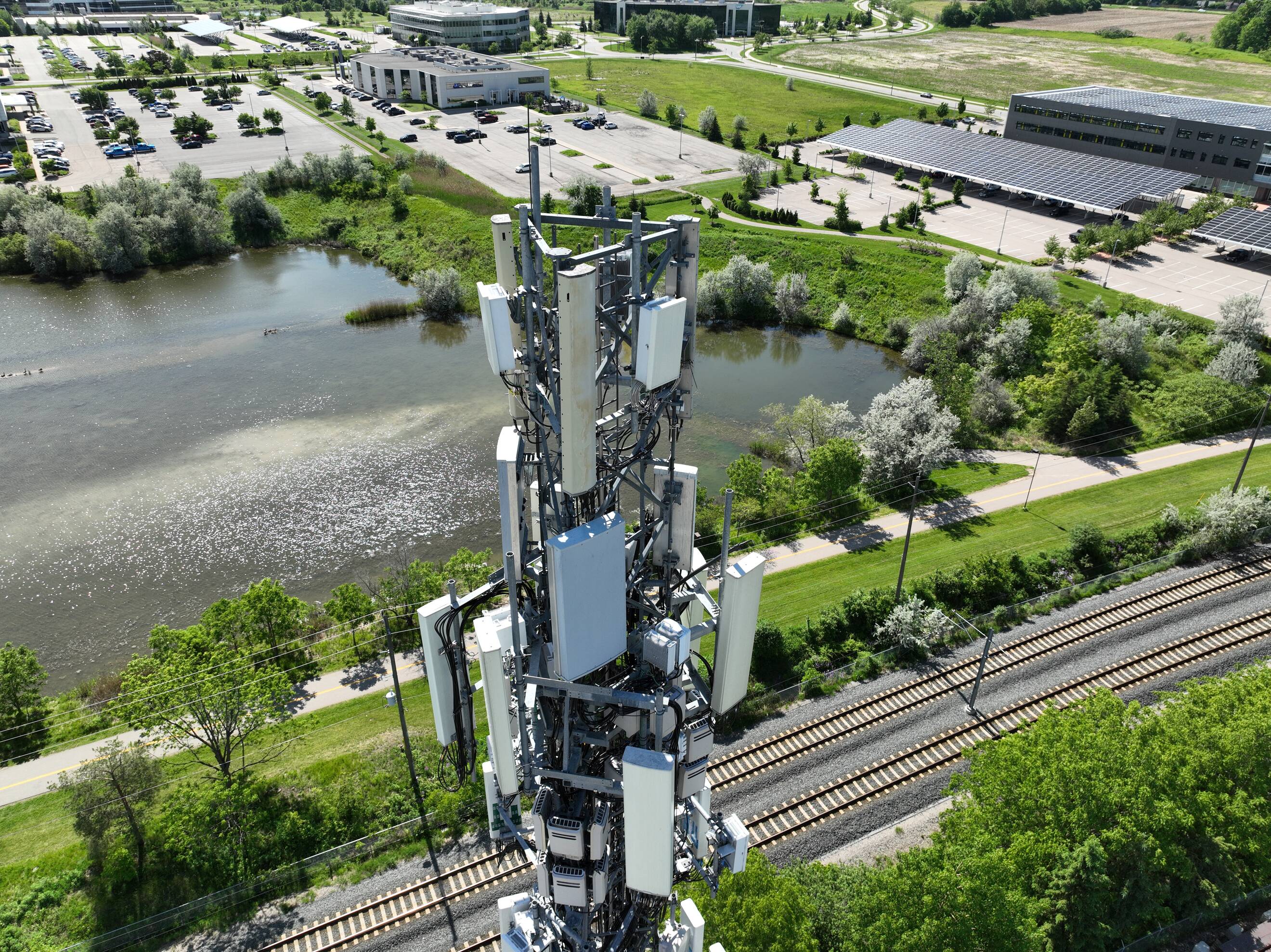} &
        \includegraphics[width=0.158\textwidth]{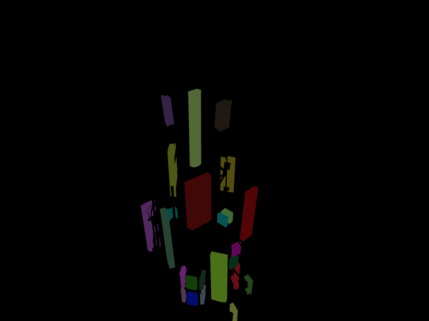} &
        \includegraphics[width=0.158\textwidth]{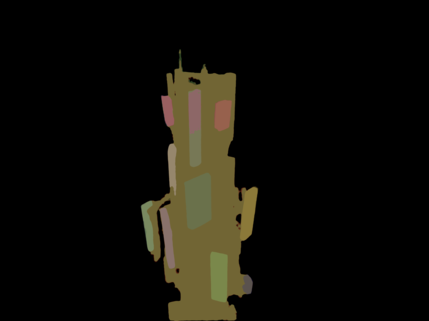} &
        \includegraphics[width=0.158\textwidth]{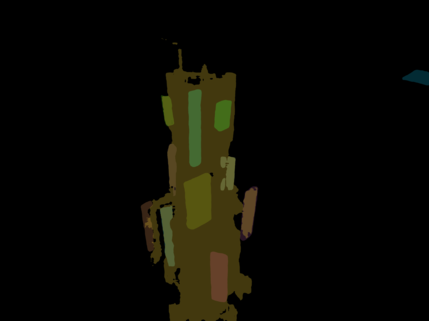} &
        \includegraphics[width=0.158\textwidth]{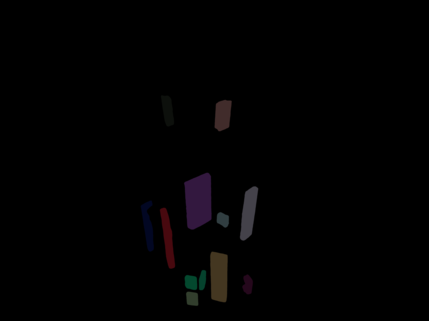} &
        \includegraphics[width=0.158\textwidth]{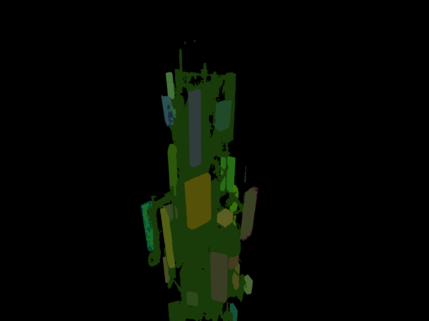} \\

        \includegraphics[width=0.158\textwidth]{figs/DJI_20240529161547_0376_V.jpg} &
        \includegraphics[width=0.158\textwidth]{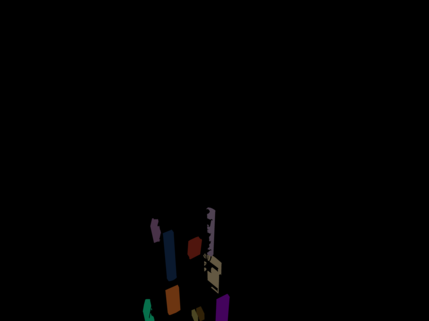} &
        \includegraphics[width=0.158\textwidth]{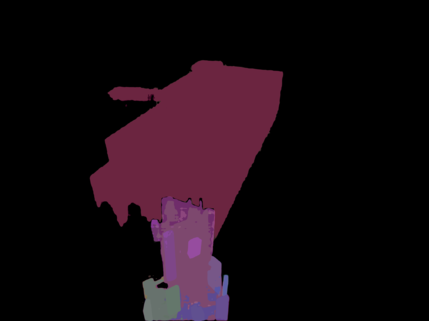} &
        \includegraphics[width=0.158\textwidth]{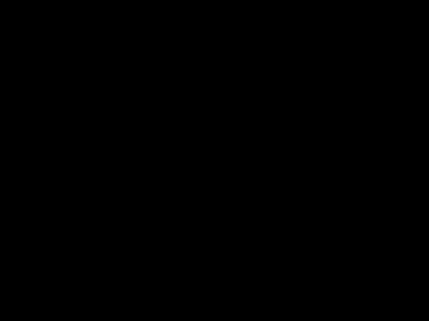} &
        \includegraphics[width=0.158\textwidth]{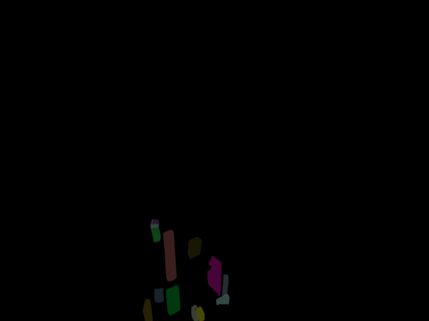} &
        \includegraphics[width=0.158\textwidth]{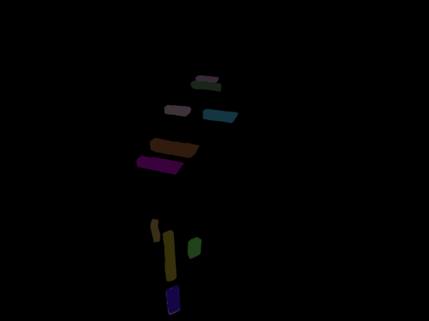} \\

        \includegraphics[width=0.158\textwidth]{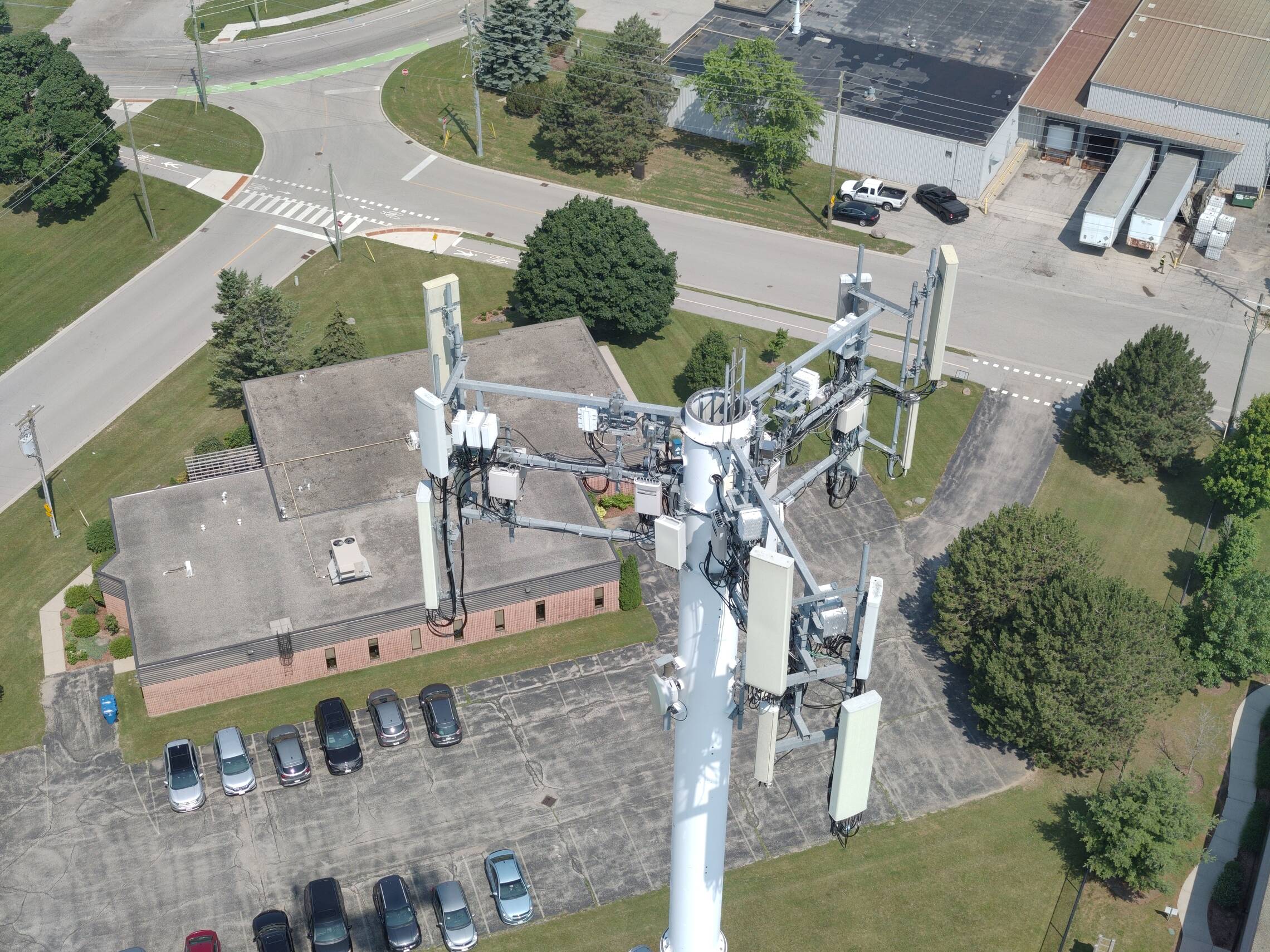} &
        \includegraphics[width=0.158\textwidth]{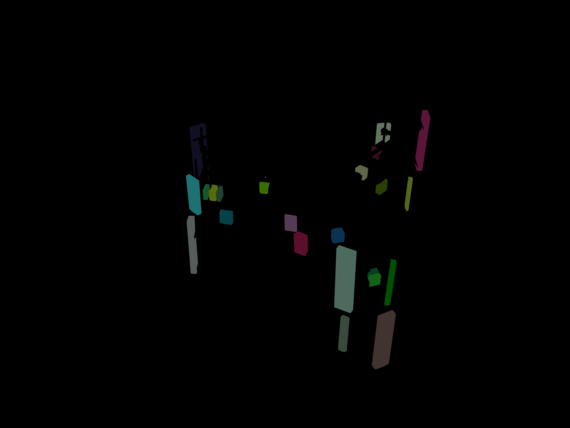} &
        \includegraphics[width=0.158\textwidth]{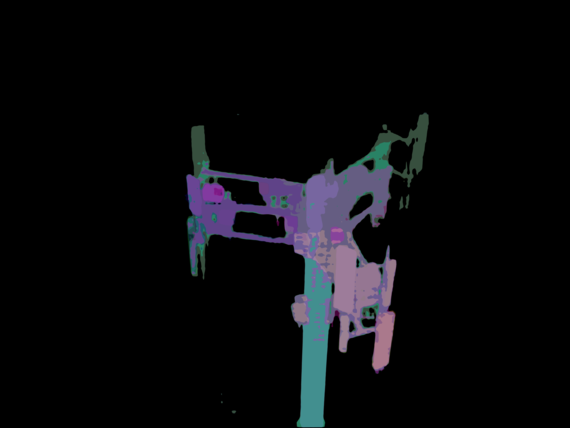} &
        \includegraphics[width=0.158\textwidth]{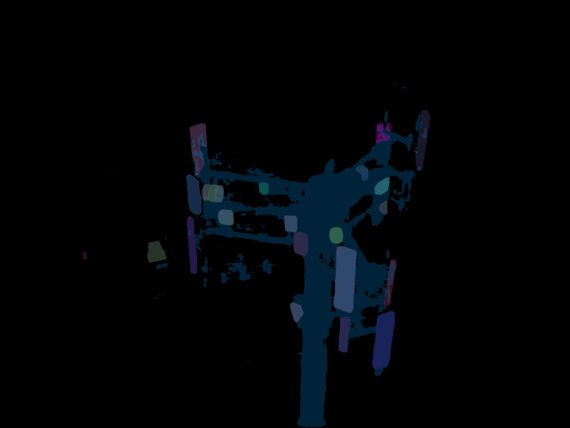} &
        \includegraphics[width=0.158\textwidth]{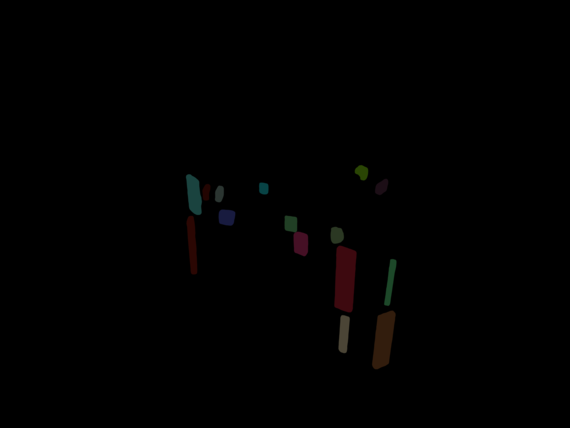} &
        \includegraphics[width=0.158\textwidth]{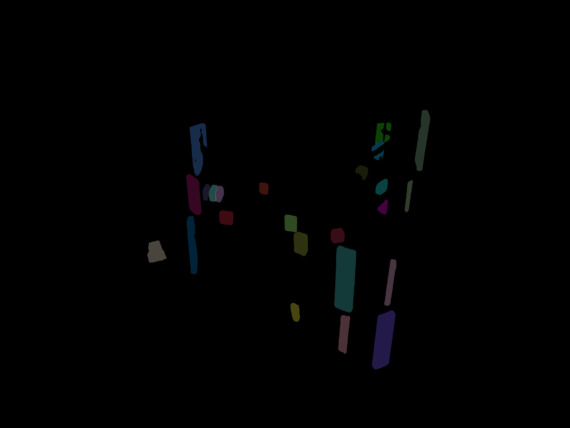}

    \end{tabular}

    \caption{\textbf{Representative failure cases.} In the first and second rows, SD-SAM~3 detects several valid components but also produces an oversized tower-level mask, whereas SD-Grounded-SAM suppresses this prediction through box refinement. The third and fourth rows show near top-down views in which foreground conditioning retains visually similar background regions, leading to localization errors.}
    \label{fig:failure-cases}
\end{figure}

Figures~\ref{fig:qualitative-results} and~\ref{fig:failure-cases} provide qualitative comparisons of the original and conditioned segmentation frameworks. Different colors are used to distinguish individual instances belonging to the same component class.

\paragraph{Effect of Saliency--Depth Conditioning.}
Figure~\ref{fig:qualitative-results} shows that saliency--depth conditioning improves both Grounded-SAM and SAM~3. The original Grounded-SAM frequently produces broad masks covering large portions of the tower and does not accurately separate individual components. SD-Grounded-SAM substantially reduces these oversized predictions and produces more compact masks that better match the annotated component regions. Similarly, SAM~3 generally localizes more instances than Grounded-SAM, but it may also respond to visually similar background objects, as observed in the second and fourth rows. SD-SAM~3 reduces much of this background confusion while preserving SAM~3's strong instance-localization capability. These visual improvements are consistent with the quantitative results, where both conditioned variants outperform their corresponding unconditioned baselines.

\paragraph{Failure Cases.}
Figure~\ref{fig:failure-cases} presents two representative challenging cases. In the first and second rows, SD-SAM~3 correctly localizes several component instances but also produces an oversized mask covering a substantial portion of the tower. When all predicted instances are merged into a binary foreground mask, this prediction introduces considerable non-target area and reduces image-level IoU and Dice. In contrast, SD-Grounded-SAM suppresses the tower-scale detection through its box-refinement stage and produces a cleaner result. This example supports the quantitative finding that SD-SAM~3 achieves higher recall and instance-level AP, whereas SD-Grounded-SAM achieves higher precision and semantic foreground quality.

The third and fourth rows illustrate the difficulty of near top-down viewpoints. From this perspective, the tower overlaps visually with rooftops and other surrounding structures, making it difficult for saliency to separate the foreground accurately. As a result, some background content can remain in the conditioned image and influence the final predictions. SD-SAM~3 recovers several plausible component instances but also includes predictions associated with the retained background. SD-Grounded-SAM again produces a more selective result because its geometric and depth-based box refinement removes many unreliable detections, although this conservative behavior may also cause some valid components to be missed.

Overall, the qualitative comparisons confirm that saliency--depth conditioning reduces background ambiguity in both frameworks, while the downstream localization and refinement mechanisms determine the final balance between instance coverage and false-positive suppression.

\subsection{Ablation Study}

\begin{table}[t]
  \centering
  \caption{Ablation study of saliency--depth conditioning and box refinement across the Grounded-SAM and SAM~3 frameworks. Best and second-best results are identified separately within each framework and are shown in bold and underlined, respectively.}
  \resizebox{\linewidth}{!}{
  \begin{tabular}{llccccc}
    \toprule
    Model
    & Conditioning
    & Box Refinement
    & mAP ($\uparrow$)
    & Precision@0.5 ($\uparrow$)
    & Recall@0.5 ($\uparrow$)
    & Semantic IoU ($\uparrow$) \\
    \midrule

    Grounded-SAM
    & None
    & \ding{51}
    & 0.2870
    & 0.8787
    & 0.3920
    & 0.6195 \\

    Grounded-SAM
    & Saliency
    & \ding{51}
    & \underline{0.3355}
    & \underline{0.8845}
    & 0.4511
    & \textbf{0.6420} \\

    Grounded-SAM
    & Saliency + depth
    & \ding{55}
    & 0.2569
    & 0.5499
    & \textbf{0.5226}
    & 0.2920 \\

    Grounded-SAM
    & Saliency + depth
    & \ding{51}
    & \textbf{0.3780}
    & \textbf{0.8970}
    & \underline{0.5073}
    & \underline{0.6281} \\

    \midrule

    SAM~3
    & None
    & -- 
    & 0.4250
    & 0.7662
    & 0.6089
    & 0.3630 \\

    SAM~3
    & Saliency
    & --
    & \underline{0.5681}
    & \underline{0.7815}
    & \underline{0.7281}
    & \textbf{0.4420} \\

    SAM~3
    & Saliency + depth
    & --
    & \textbf{0.5701}
    & \textbf{0.7817}
    & \textbf{0.7311}
    & \underline{0.4376} \\
    
    \bottomrule
  \end{tabular}
  }
  \label{tab:ablation}
\end{table}

Table~\ref{tab:ablation} evaluates the individual contributions of saliency conditioning, depth-based recovery, and box refinement across Grounded-SAM and SAM~3. The shared conditioning components are progressively introduced in both frameworks, while paired Grounded-SAM configurations with and without box refinement isolate the effect of the additional prompt-filtering stage. We report mAP, precision, recall, and semantic IoU to capture instance discovery, false-positive suppression, and aggregate foreground quality.

\paragraph{Effect of Foreground Conditioning.}
For Grounded-SAM, saliency increases mAP from 0.2870 to 0.3355 and recall from 0.3920 to 0.4511 when box refinement is enabled. Adding depth further increases mAP to 0.3780 and recall to 0.5073. A similar trend is observed for SAM~3, where saliency increases mAP from 0.4250 to 0.5681 and recall from 0.6089 to 0.7281, while adding depth further improves them to 0.5701 and 0.7311, respectively. These results indicate that saliency reduces background ambiguity, whereas depth helps recover additional near-field component regions that may be missed by saliency alone. The small decrease in semantic IoU after adding depth suggests that this higher instance coverage can also retain some nearby non-target regions.

\paragraph{Effect of Box Refinement.}
The importance of box refinement is evident when comparing the two saliency--depth-conditioned Grounded-SAM configurations. Without box refinement, recall reaches 0.5226, slightly higher than the 0.5073 obtained by the complete pipeline. However, precision decreases sharply from 0.8970 to 0.5499, mAP falls from 0.3780 to 0.2569, and semantic IoU drops from 0.6281 to 0.2920. This shows that the refinement stage effectively removes oversized, redundant, and background detections, producing a cleaner set of spatial prompts with only a small loss in recall.

\paragraph{Overall Findings.}
Overall, the ablation study shows that saliency is the primary source of improvement across both segmentation frameworks, while depth complements it by recovering additional near-field regions and improving instance discovery. Box refinement suppresses unreliable detections, thereby substantially increasing precision and aggregate foreground quality. Together, these components provide complementary benefits: saliency--depth conditioning improves component discovery across both frameworks, while the Grounded-SAM-specific refinement stage improves the reliability of the final predictions.

% //////////////////////////////////////////////////////////////////
% /////////////////////////  Conclusion  ///////////////////////////
% //////////////////////////////////////////////////////////////////

\section{Conclusion}
This work introduced a model-agnostic saliency--depth foreground-conditioning strategy for training-free segmentation of communication-tower components in cluttered UAV imagery. The proposed conditioning module combines appearance-based saliency with monocular relative depth to suppress irrelevant background regions while recovering near-field tower structures that may be missed by saliency alone. We integrated this strategy with two architecturally distinct zero-shot segmentation frameworks, resulting in SD-Grounded-SAM and SD-SAM~3. For the Grounded-SAM branch, we additionally introduced a geometric and depth-aware box-refinement stage to improve the reliability of the spatial prompts provided to SAM. Experiments on the author-constructed TOW-300 dataset demonstrate that the proposed conditioning strategy consistently improves both underlying frameworks across instance- and image-level segmentation metrics. SD-SAM~3 achieves the strongest overall instance segmentation performance, showing that foreground conditioning improves SAM~3's internal concept localization and component discovery. SD-Grounded-SAM achieves the highest precision and semantic IoU, reflecting the effectiveness of its explicit box-refinement stage in suppressing oversized, redundant, and background detections. The ablation study further shows that saliency reduces background ambiguity, depth improves instance discovery by recovering additional near-field regions, and box refinement substantially improves false-positive suppression and aggregate foreground quality. These findings indicate that background conditioning addresses a general limitation of zero-shot segmentation in visually complex infrastructure scenes rather than a weakness specific to a single model. Nevertheless, depth-based recovery may retain some nearby non-target structures, while conservative box filtering can remove small or weakly visible components. Future work will investigate adaptive fusion and filtering strategies, improved foreground-prior estimation, and evaluation across additional tower configurations and infrastructure datasets to further improve the balance between component discovery and false-positive suppression.

% \section*{References}
\bibliographystyle{ieeetr}
\bibliography{ref.bib}

\end{document}